\documentclass[AMA,STIX1COL]{WileyNJD-v2}

\articletype{Article Type}%

\pdfoutput=1
\usepackage{url} 
\usepackage{hyperref}
\usepackage{color}
\usepackage{ulem}
\usepackage{subfig}
\usepackage{longtable}
\usepackage{supertabular,booktabs}
\usepackage{makecell}
\usepackage{multicol,lipsum}
\usepackage{graphicx}

\begin{document}

\title{Reduced-Gate Convolutional LSTM Using Predictive Coding for Spatiotemporal Prediction}
\author[1]{Nelly Elsayed*}

\author[2]{Anthony S. Maida}

\author[2]{Magdy Bayoumi}

\authormark{AUTHOR ONE \textsc{et al}}

\address[1]{\orgdiv{School of Information Technology}, \orgname{University of Cincinnati}, \orgaddress{\state{Ohio}, \country{USA}}}

\address[2]{\orgdiv{School of Computing and Informatics}, \orgname{University of Louisiana at Lafayette}, \orgaddress{\state{Louisiana}, \country{USA}}}

\corres{*Nelly Elsayed, \email{nelly.elsayed@uc.edu}}

\abstract[Summary]{Spatiotemporal sequence prediction is an important problem in deep learning. We study next-frame(s) video prediction using a deep-learning-based predictive coding framework
	that uses convolutional, long short-term memory (convLSTM) modules.
	We introduce a novel reduced-gate convolutional LSTM (rgcLSTM) architecture that requires a significantly lower
	parameter budget than a comparable convLSTM.
	By using a single multi-function gate,
	our reduced-gate model achieves equal or better next-frame(s) prediction accuracy than the original convolutional
	LSTM while using a smaller parameter budget, thereby reducing training time and memory requirements.
	We tested our reduced gate modules within a predictive coding architecture 
	on the moving MNIST and KITTI datasets.
	We found that our reduced-gate model has a significant reduction of approximately 40 percent of 
	the total number of training parameters and a 25 percent reduction
	in elapsed training time in comparison with the standard convolutional LSTM model.
	The performance accuracy of the new model was also improved.
	This makes our model more attractive for hardware implementation, especially on small devices.
	We also explored a space of twenty different gated architectures to get insight into how our rgcLSTM fits into that space.
}


\maketitle

\footnotetext{\textbf{Abbreviations:} rgcLSTM, reduced-gate convolutional LSTM; LSTM, long short-term memory}

\section{Introduction}
The brain in part acquires representations by using learning
mechanisms that are triggered by prediction errors 
when processing sensory input~\cite{Friston2005,Bastos2012a,rao1999predictive}. An early implementation using this approach is by Rao et al.~\citep{rao1999predictive}
to model
non-classical receptive field properties in the neocortex.
The hypothesized brain mechanisms underlying this ``predictive coding'' are based on the concept of
bi-directional interactions between the higher and lower-level areas of the visual cortex.
The higher-level areas send predictions about the incoming sensory input to the
lower-level areas. 
The lower-level areas compare the predictions with ground truth sensory input and calculate the prediction errors.
These are in turn forwarded to the higher-level areas to update their predictive representations
in light of the new input information.

Lotter et al.~\citep{Lotter2017} introduced a predictive coding architecture
to the deep learning community.
Specifically, they introduced PredNet for next-frame 
video prediction.
The architecture was based on a deep neural-network framework that used a hierarchy of 
convolutional
LSTMs.
Since it was a deep network, it could readily be implemented and studied using off-the-shelf deep learning 
frameworks (e.g., Keras~\citep{kerasAPI}, PyTorch~\citep{paszke2017automatic}, TensorFlow~\citep{tensorflow2015-whitepaper}).
This was in contrast to earlier 
predictive coding
models~\citep{Friston2005} with a 
primarily
mathematical formulation
that offered little or no implementation guidance.

One of the main module types within the PredNet architecture is the representation module which is implemented
as a basic convolutional LSTM (cLSTM).
The module, when coupled with other modules, supports gated recurrent processing of video sequences
and is trained to anticipate, or predict, the next frame that will appear in the sequence.

Our work replaces the cLSTM modules within the Lotter at al.~\citep{Lotter2017} model with a gated convolutional network
that has fewer parameters.
Its distinguishing feature is that it uses a single gate to perform three different functions.
We call this the reduced-gate convolutional network, or ``rgcLSTM''
which introduces the concept of multi-function gates. 
The single gate uses shared weights to perform three functions and we classify it as a \textit{multi-function gate model}.
This allows a gate to perform more than one traditional function, thereby reducing
the number of needed gates and the associated parameter count.
Because of its smaller parameter budget, the rgcLSTM has a smaller memory footprint and faster training while maintaining
or improving the prediction performance accuracy of the original model.

\begin{figure}
	\centering
	\includegraphics[width=7.2cm,height=5.5cm]{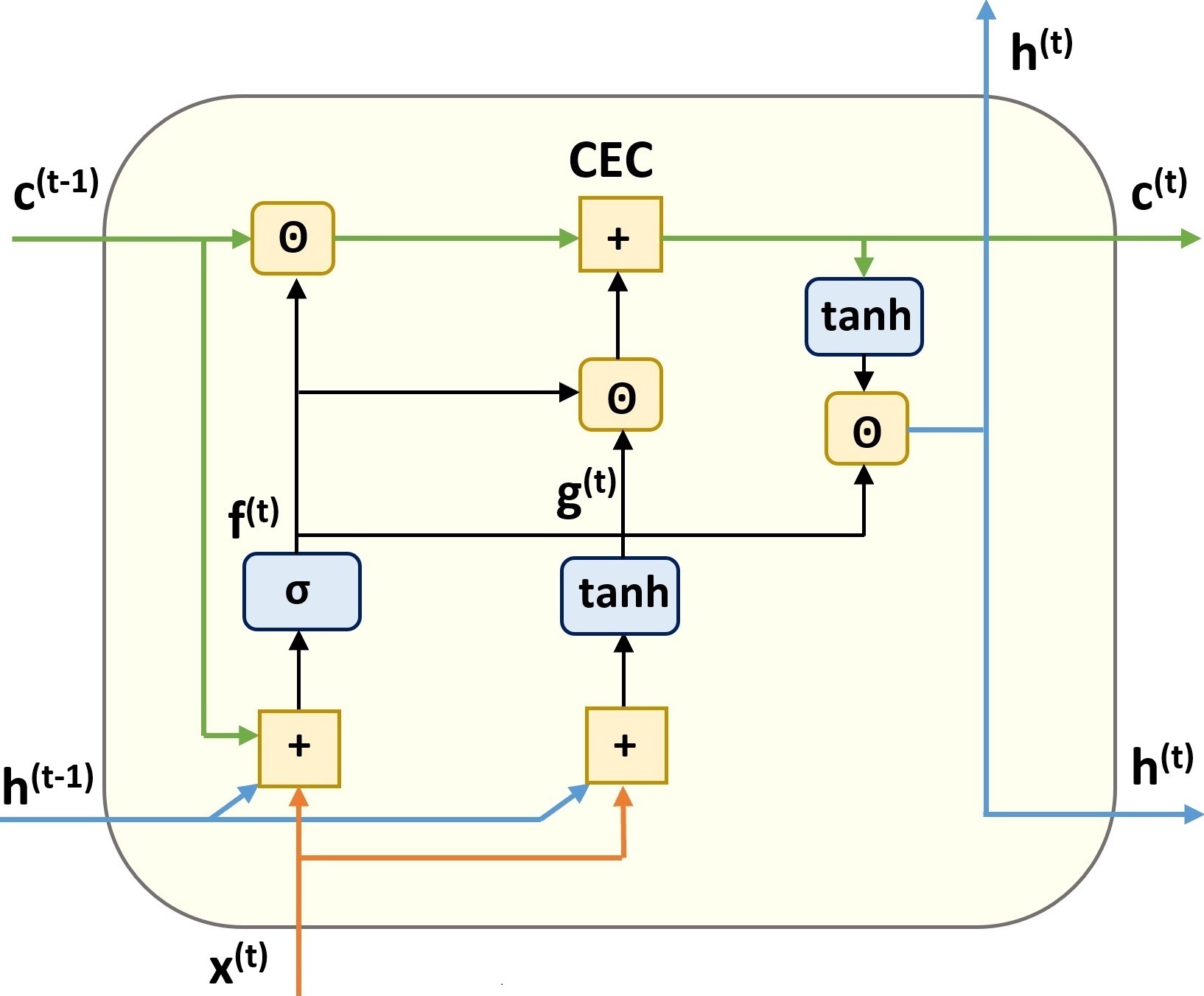}
	\caption{
		Diagram of an unrolled rgcLSTM cell.
	    The single module gate, \textbf{f}, 
		sends information to three locations which correspond to the output values of the forget, input, and output
		gates of the standard LSTM.
		}
	\label{rgcLSTM_architecture}
\end{figure}

The present paper motivates
and evaluates
the design of our rgcLSTM, 
shown in Figure~\ref{rgcLSTM_architecture},
whose novel features are
the use of
convolution-based peephole connections and the single 
multi-function 
gate 
that serves the function of three gates, namely, the forget gate, the input gate, and the output gate.
We present performance results on the moving MNIST (gray-scale) and KITTI 
traffic
(RGB) dataset benchmarks with, and without, these modifications. 
We find that our architecture gives better next-frame(s) prediction 
accuracy
on these datasets while using a smaller parameter budget
in comparison to the original Shi et al.~\citep{Shi2015} convLSTM.
We also explore a space of twenty different gated RNN designs
involving both multi-function and single-function gates.
This enriches our insight into how these gated networks perform.

\section{Background and Related Work}
Recurrent neural networks (RNNs) 
process sequential data such as 
occurs in
signal processing~\citep{sohn2001damage}, weather feeds~\citep{rotton1985air}, time series~\citep{elsayed2018deep}, and videos~\citep{Lotter2017,Shi2015}.
Spatiotemporal datasets such as video are sequential datasets where 
the sequence elements are images.
Spatiotemporal prediction (e.g., video prediction) is a challenge that has received 
intense interest in deep learning over the last few years. 
Spatiotemporal prediction and video frame prediction typically use unsupervised learning~\citep{srivastava2015unsupervised,Shi2015,Lotter2017,finn2016unsupervised}.
However, the models all use 
complicated
architectures and a large parameter budget.

The module that we modify uses convolutional LSTM modules.
LSTM modules are recurrent neural networks (RNNs) which have three gating mechanisms.
Plain RNNs do not have gating mechanisms and suffer from vanishing and/or exploding
gradient problems.
Although, they can learn local sequential dependencies, the gradient issues prevent them from
learning long-term dependencies.

\begin{figure*}
	\centering
		\subfloat[]{\includegraphics[width=0.375\textwidth, height = 5.0cm]{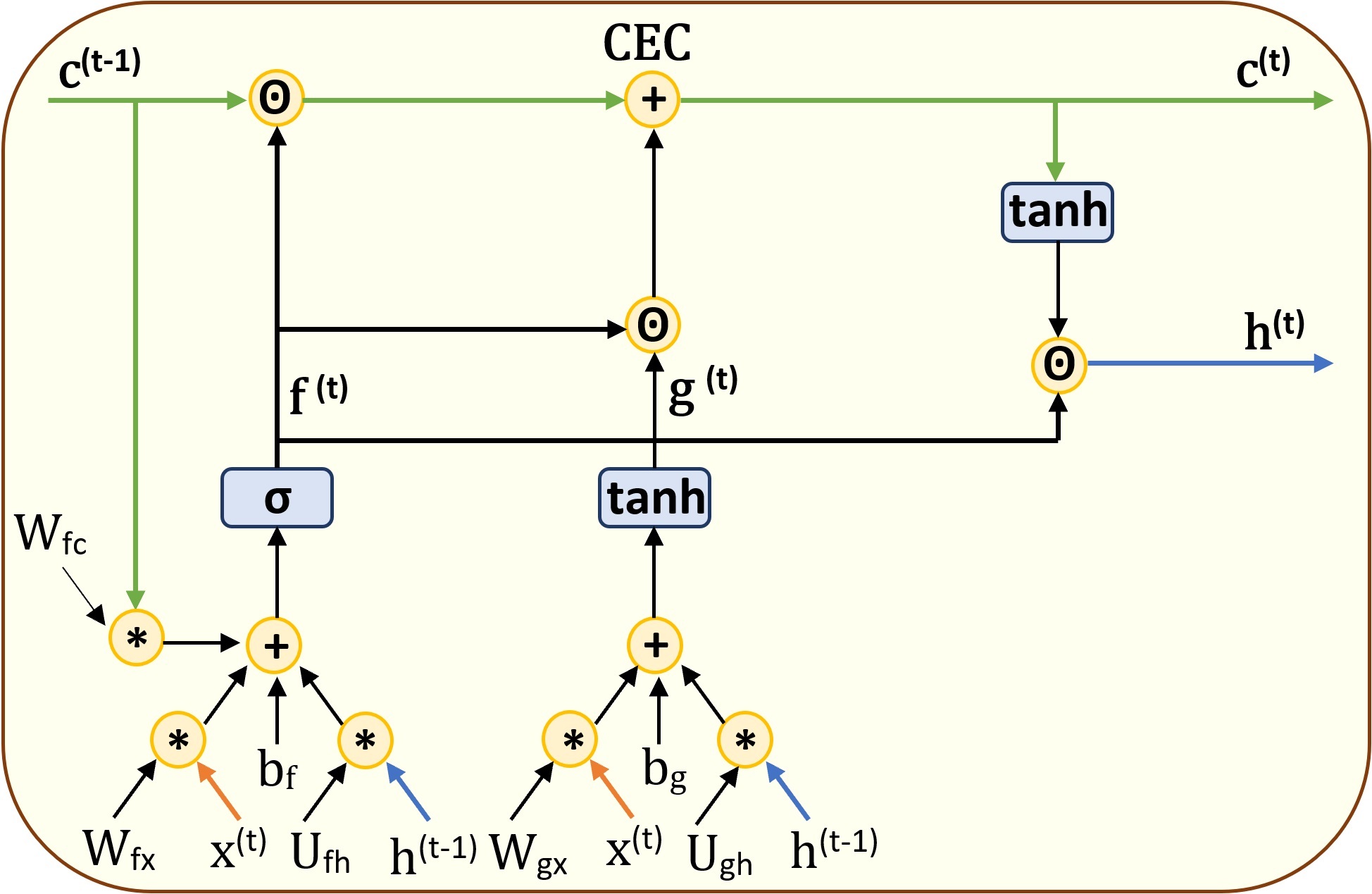}}
		\qquad\qquad
		\subfloat[]{\includegraphics[width=0.417\textwidth, height= 5.0cm]{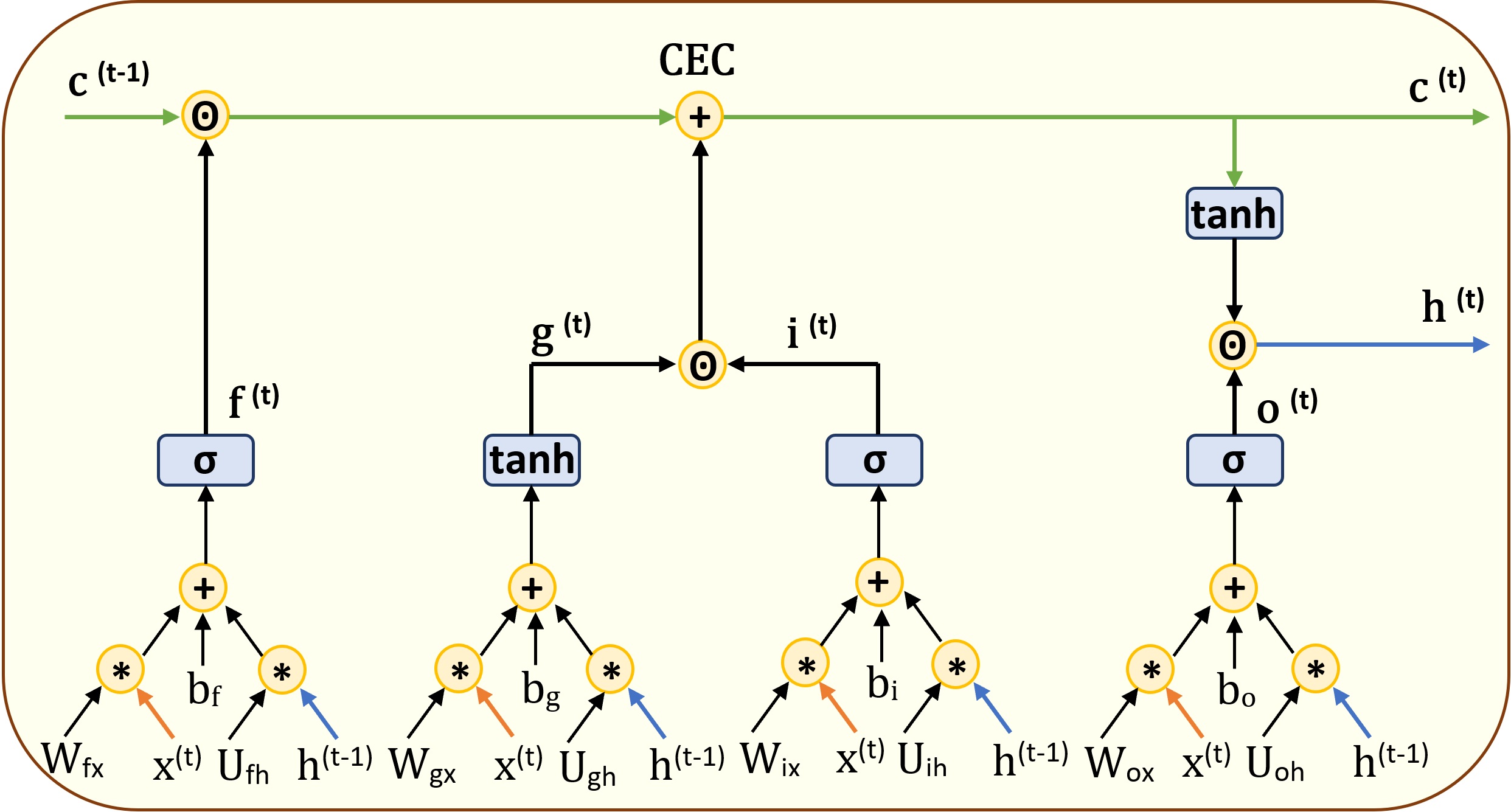}}
		\caption{An operational diagram to make the trainable weights and biases explicit, along with their arithmetic operations, for the (a) rgcLSTM;
		and (b) cLSTM cell architectures. The proposed rgcLSTM has approximately 40\% fewer trainable parameters than the cLSTM.}
		\label{rgcLSTM_weights}

\end{figure*}

The long short-term memory (LSTM) network,
first 
introduced
by Hochreiter and Schmidhuber~\citep{hochreiter1997a}, was the first gated RNN
recurrent unit 
design
to mitigate the vanishing and/or 
exploding gradient problems
and improve learning of long-term dependencies.
The LSTM was a recurrent block, or cell, with considerable internal structure
similar to that in Figure~\ref{rgcLSTM_weights}(b), which displays a convolutional LSTM cell.
Modern LSTMs have three gates and an input-update assembly.
The gate names are ``forget,'' ``input'', and ``output''.
To support learning, these gates employ trainable weights which are affine multiplied
with the relevant gate input and then added to a trainable bias.~\citep{greff2017lstm}
This lets the data streams within the cell learn to remember or forget past information
that is relevant or irrelevant to the task at hand for the current point in the data stream.
Hochreiter and Schmidhuber~\citep{hochreiter1997a} introduced a second innovation
to specifically address the vanishing/exploding gradient problems.
They replaced the multiplicative transfer to previous states by additive transfer
(indicated by the plus in the $c$ line of Figure~\ref{rgcLSTM_architecture}).
The component to implement this was called the constant error carousel (CEC).
Although mitigating the vanishing/exploding gradient problems, the CEC introduced a weakness into
the LSTM design that became apparent when the output gate was closed.
In this situation, the CEC could no longer affect the function of the input and forget gates
(because the $h$ inputs are cut off, as seen in Figure~\ref{rgcLSTM_weights}(b)).
This problem was handled by 
adding peephole connections 
from the memory cell-state data line
to each of the LSTM gates. This, however, comes at a cost of increasing the number of trainable
parameters and associated overhead such as training time and memory footprint.
For the rgcLSTM, only one peephole connection is needed.
It goes from the cell state to the net input of the forget gate (see Figure~\ref{rgcLSTM_architecture}
and Figure~\ref{rgcLSTM_weights}(a)).

There were attempts to design gated models to reduce the gate count while
preserving learning power. 
Cho et al.~\citep{cho2014properties} proposed a gated recurrent unit (GRU) model. 
Instead of three gates, it used two: an update gate and a reset gate. 
The update gate combined the input gate, forget gate~\citep{greff2017lstm}. 
Thus, this is the first multi-function gate model.
The reset gate 
served
as the output gate of the LSTM cell~\citep{greff2017lstm}. 
This GRU model eliminated the output activation function and memory unit, but retained CEC\@~\citep{greff2017lstm}. 
The GRU yielded a reduction in trainable parameters compared with the standard LSTM\@.
Zhou et al.~\citep{zhou2016minimal} used a single gate recurrent model called a minimal gated unit (MGU)\@. 
Both models reduced the number of trainable parameters and gave results comparable to the LSTM~\citep{chung2014empirical}. 
There was a study based on the LSTM model in~\citep{gers2000learning} to examine the role and significance of each gate in the LSTM~\citep{heck2017simplified,jozefowicz2015empirical}. This study showed that the most significant gate in the LSTM was the forget gate.
However, it also showed that the forget gate needs other support to enhance its performance.

While the above gated RNNs are designed to process
non-spatial
sequential data, they must be augmented somehow
to process 
spatiotemporal
data.
Shi et al.~\citep{Shi2015} proposed a convolutional LSTM to enhance performance for spatiotemporal prediction,
which we henceforth call the convLSTM to distinguish between the Lotter et al.~\citep{Lotter2017} cLSTM\@.
The convLSTM operated on sequences of images.
It had three gates characteristic of the regular LSTM, as well as elementwise peephole connections.
The cLSTM was the same as the convLSTM, but without the peephole connections.
Both models replaced the matrix multiplications (affine transformations)
by convolutional operations for the input and recurrent input of each gate.
The model achieved higher accuracy in comparison to the classical LSTM model.
However, the number of parameters remained higher because the peephole connections still used elementwise  multiplication. 
In recent research, Lotter et al.~\citep{Lotter2017} showed how to build the first predictive coding model using a 
simpler version of the convLSTM architecture which we call the cLSTM\@.
The cLSTM had three gates but no peephole connections.
The model achieved significant improvement in a predictive model as it did not need an encoder and decoder~\citep{Lotter2017}.
However, the number of parameters was large and grew linearly with the number of layers in the model.

There were other attempts to design smaller recurrent gated models based on the standard LSTM. 
They were based on either removing one of the gates or the activation functions from the standard LSTM unit. 
Empirical analysis~\citep{greff2017lstm} compared these models and the standard LSTM. 
The conclusion found in Greff et al.~\citep{greff2017lstm} was that these models had no significant improvement
in either performance or training time. 
Moreover, these models had no significant reduction in trainable parameters. 
This empirical analysis~\citep{greff2017lstm} also stated that the critical components of the LSTM model were the forget gate and the output activation function. 

Our new gated model is named the \textit{reduced-gate, convolutional LSTM} (rgcLSTM).
Based on the empirical results of Greff et al.~\citep{greff2017lstm}, our model 
preserves the critical components of the LSTM while removing parameter redundancy.
We use our model within a predictive coding framework introduced by Lotter et al.~\citep{Lotter2017} as a state-of-the-art approach for spatiotemporal prediction.
The Lotter et al.~\citep{Lotter2017} model uses standard convLSTM modules (no peephole connections) within their predictive coding architecture.
We replace those modules with 
our
rgcLSTM.
Our model shows 
a prediction accuracy
comparable to 
the
Lotter et al.~\citep{Lotter2017} design. 
However, our rgcLSTM reduces the number of trainable parameters and memory requirements by about $40\%$ and 
the training time by $25\%$ which makes it more attractive
for hardware implementation
on low power devices.

\section{Reduced-gate Convolutional LSTM Architecture}
\label{sec:rgcLSTM}

The proposed rgcLSTM 
cell design
appears in 
Figures~\ref{rgcLSTM_architecture} and \ref{rgcLSTM_weights}(a).
Figure~\ref{rgcLSTM_weights} 
compares
the proposed rgcLSTM and 
cLSTM 
architectures in 
enough detail for meaningful comparison of the
arithmetic operations.

The
rgcLSTM
architecture has one trainable gated unit which we 
call the forget gate or module gate. 
Because of its inputs, it is 
comparable to an LSTM
forget gate, but since it is the only gate in the module it also
makes sense to call it the module gate.
The model also preserves the cell memory state and 
retains
the CEC to avoid 
vanishing and/or exploding gradients. 
There is a peephole connection from the cell state to the module gate but we have converted its operator from 
elementwise multiplication
to convolution.
This 
reduces the learning capacity of the rgcLSTM in comparison to a full LSTM with elementwise peephole connections
to 
projecting
all three gates.
However, it 
still
preserves information needed to allow the memory state to exert control over the module gate.
It also helps
the model 
extract features.
Also, our model retains the output activation function. 
Thus, our model preserves the critical components of the LSTM as stated by Greff et al.~\citep{greff2017lstm} while removing much of the parameter redundancy in the LSTM unit. 
This 
provides
a significant reduction in the 
needed
number of 
trainable parameters.
Also,
experiments in this paper show that
our rgcLSTM model preserves the
Lotter et al.~\citep{Lotter2017}
model's prediction accuracy results.

The following describes the operation of the rgcLSTM depicted in Figure~\ref{rgcLSTM_weights}(a).
We start by characterizing the forget gate.
$a^{t}$ denotes the net input to the forget/module gate activation function $\sigma(\cdot)$.
Its value is calculated using the formula below.

\begin{equation}\label{main_eqn}
a^{(t)} = \left[ W_{fx}, U_{fh}, W_{fc} \right] * \left[x^{(t)}, h^{(t-1)}, c^{(t-1)}\right] + b_f
\end{equation}

\noindent
$a^{(t)}$ is an $\eta\times\upsilon$ image with $n$ channels.
The square brackets on the righthand side of the equation are a channel-stacking operator
and the ``*'' symbol denotes `same' convolution.
$x^{(t)}$ denotes the current multi-channel image input to the module.
Its dimensions are $\eta\times\upsilon$ with $\gamma$ channels.
$h^{(t-1)}$ denotes the output of the cell from the previous time step and
$c^{(t-1)}$ denotes the cell state from the previous time step.
$c$ and $h$ have the same dimensions and number of channels as $a$.
$W_{fx}$, $U_{fh}$, and $W_{fc}$ are $m\times m$ convolution kernels to go with
the cell inputs $x^{(t)}$, $h^{(t-1)}$, and $c^{(t-1)}$.
$W_{fx}$ has $\gamma$ channels and $U_{fh}$ and $W_{fc}$ each have $\kappa$ channels.
$W_{fc}$ implements the convolutional peephole connection.
To match the output channel size for $a$, the following constraint is enforced:
$\upsilon=\gamma + 2\kappa$.
All three weight sets and biases $b_f$ are trainable.
To condense the notation we let $W_f=[W_{fx}, U_{fh}, W_{fc}]$ and
$I_f = [x{(t)}, h^{(t-1)}, c^{(t-1)}]$.
Bias $b_f$ is an $n$-element vector and is added to the appropriate image channel by broadcasting.

Note that the convolution operation between $W_{fc}$ and $c^{(t-1)}$ 
implements the peephole connection and
represents a departure from
Shi et al.~\citep{Shi2015} where an 
elementwise multiply 
was used 
(in contrast to 
a
convolution
operation in our model
).

The total number of trainable parameters for the module gate is:

\begin{equation}
\label{eqnFgateParamCount}
f_\mathrm{gate}^\# = \left(m^2(\gamma+2\kappa)+1\right)\cdot n.
\end{equation}

\noindent
where the superscript ``\#'' is a tag to indicate number of parameters.
$m^2$ is the number of filter weights.
$\gamma+2\kappa$ is the number of input channels. 
$n$ is the number of filters.

The module gate image value, $f_{gate}^{(t)} \in \mathbb{R}^{\eta\times\upsilon\times n}$, 
is obtained by applying a pixelwise activation function $G$ 
to the net input image using
\begin{equation}
\label{eqn_f_gate}
f_\mathrm{gate}^{(t)}= G(a^{(t)}) .
\end{equation}
Depending on the application, $G$ can be either the logistic ($\sigma$) or hard sigmoid ($hardSig$) function~\citep{Gulcehre2016}. 
The pixel values of $f^{(t)}$
will fall in the range $(0,1)$ or $\left[0, 1\right]$, 
depending on which function is used.
Using $\sigma$, the gate value $f^{{(t)}}$ is calculated by:

\begin{equation}
f^{(t)} = \sigma(W_{f}*I_f + b_{f}) .
\label{forget_gate_eqn}
\end{equation}

\noindent
Stacking
makes the learning process more powerful than the non-stacked weights due to the influence of the $x^{t}$, $h^{(t-1)}$ and $c^{(t)}$ 
across convolutional weight set. 
Lotter et al.~\citep{Lotter2017} and Heck et al.~\citep{heck2017simplified} show empirically
that stacking the input for recurrent units achieves better results than non-stacked input. 

The input update (memory activation) uses a similar equation as for the module gate.

\begin{equation}
\label{memoryActEqn}
g^{(t)} =\tanh\left(W_g * I_g + b_g\right)
\end{equation}

\noindent
In the above, $W_g = \left[ W_{gx}, U_{gh} \right] \in \mathbb{R}^{m  \times m \times\left(\gamma+\kappa\right) \times n}$, and $I_g = \left[x^{(t)}, h^{(t-1)}\right] \in \mathbb{R}^{\eta  \times \upsilon \times\left(\gamma+\kappa\right)}$.
The number of channels in $W_g$ is matching the number of 
channels of the $W_f$ which make them computationally compatible. This approach is taken so that the dimension of $g_\mathrm{gate}^{(t)} \in \mathbb{R}^{\eta\times\upsilon\times n}$
matches the dimension of $f_\mathrm{gate}^{(t)}$.
Similarly, the dimension of $b_g \in \mathbb{R}^{n\times 1}$ matches that of $b_f \in \mathbb{R}^{n\times 1}$.
Finally, the number of trainable parameters for the input update is:

\begin{equation}
\label{eqnInputUpdateParamCount}
g_\mathrm{update}^\# = \left(m^2(\gamma+\kappa)+1\right)\cdot n.
\end{equation}

\noindent
Eqns.~\ref{eqnFgateParamCount} and~\ref{eqnInputUpdateParamCount} count the total number of trainable parameters for the rgcLSTM
cell.
Eqn.~\ref{eqnInputUpdateParamCount} differs from Eqn~\ref{eqnFgateParamCount}
because the former does not have peephole connections.
The 
total rgcLSTM parameter 
count is 

\begin{equation}
\label{parameters_count}
\mathrm{rgcLSTM}_\mathrm{cell}^\# = f_\mathrm{gate}^\# + g_\mathrm{update}^\# = 2\cdot\left(m^2(\gamma+\frac{3}{2}\kappa)+1\right)\cdot n
\end{equation}

The final equations to complete the specification of the rgcLSTM are given below.

\begin{align}
c^{(t)} &= f^{(t)} \odot c^{(t-1)} + f^{(t)} \odot g^{(t)}\label{memory_cell}\\
h^{(t)} &= f^{(t)} \odot tanh(c^{(t)})\label{rgcLSTM_output_eqn}
\end{align}

\noindent
The $\odot$ symbol denotes
elementwise multiplication. 
$\kappa$ is constrained to equal $n$ so that the dimensions match for the elementwise multiplication operations.
Eqn.~\ref{memory_cell} uses a ``+'' operator on the righthand side and this implements the CEC\@.

Table~\ref{components_table} 
compares features
between
our
proposed rgcLSTM
versus the
cLSTM and convLSTM 
designs.
Also, model M18 in Table~\ref{models_shared} of the appendix is an alias for rgcLSTM and
gives the summary equations for our rgcLSTM model.
Similarly, models M1 and M8 are aliases for the cLSTM and convLSTM, respectively, and their summary equations
are described in Table~\ref{non_shared_models} of the appendix.

\begin{table*}[t]%
\begin{center}
		\caption{
			     Feature
		         comparison between rgcLSTM, cLSTM and convLSTM 
		         cells.
		         \label{components_table}
	         }
		\centering
		\begin{tabular*}{330pt}{@{\extracolsep\fill}lcccD{.}{.}{3}c@{\extracolsep\fill}}
			\toprule
			\textbf{Component} & \textbf{rgcLSTM}  & \textbf{cLSTM} &\textbf{convLSTM}\\ 
			\midrule
			Number of gates.&1&3&3\\
			Number of  
			non-gate
			activation 
			functions (tanh).
			& 2&2&2\\ 
			Has peephole connection?&Yes&No&Yes\\
			Number of convolution kernels.&5&8&8\\
			Number of 
			non-convolutional 
			weight matrices.&0&0&3\\
			Number of elementwise multiplications.&3&3&6\\
			Number of convolutional multiplications.&5&8&8\\
			Number of bias vectors.&2&4&4\\
			\bottomrule
		\end{tabular*}
\end{center}
\end{table*}

\begin{figure}
	\centering
	\includegraphics[width=8.5cm, height=6cm]{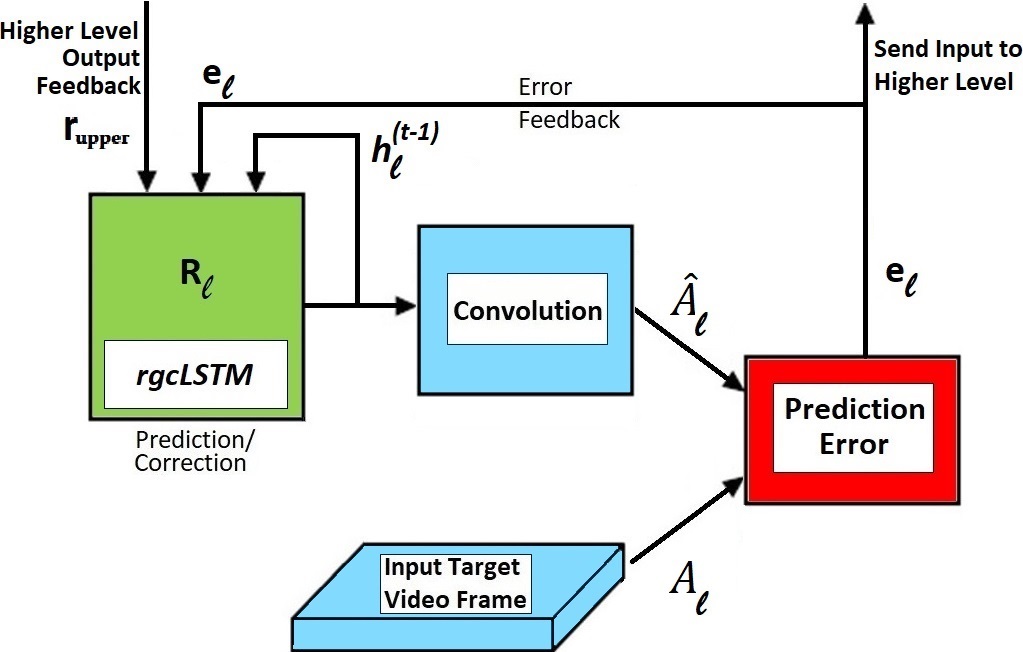}
	\caption{
		 The lowest
	     layer of the PredNet architecture with our rgcLSTM module substituted for the cLSTM 
	     used
	     in the original PredNet~\citep{Lotter2017}. 
	     The layer subscript $l$ is used to signify that higher layers obey the
		same design, with the caveat that the input to higher layers is prediction error from the 
		immediately
		lower layer.}
	\label{figOverview}
\end{figure}

\section{Overall Architecture}

We use our rgcLSTM module within the PredNet predictive coding framework of Lotter et al.~\citep{Lotter2017}, 
so have named our variant 
Pred-rgcLSTM~\citep{ijcnnrgc}.
The bottom
layer of the architecture appears in Figure~\ref{figOverview}. 
Our contribution lies 
in the redesign of the 
R module
where our rgcLSTM replaces the 
cLSTM 
used by Lotter et al.
In this design, 
whether PredNet or Pred-rgcLSTM, learning 
primarily
occurs within the gated recurrent 
cells
but is triggered by prediction error signals
that are input to the R module.

\begin{figure*}
	\centering
	\includegraphics[keepaspectratio=true,width=\textwidth,height=20cm]{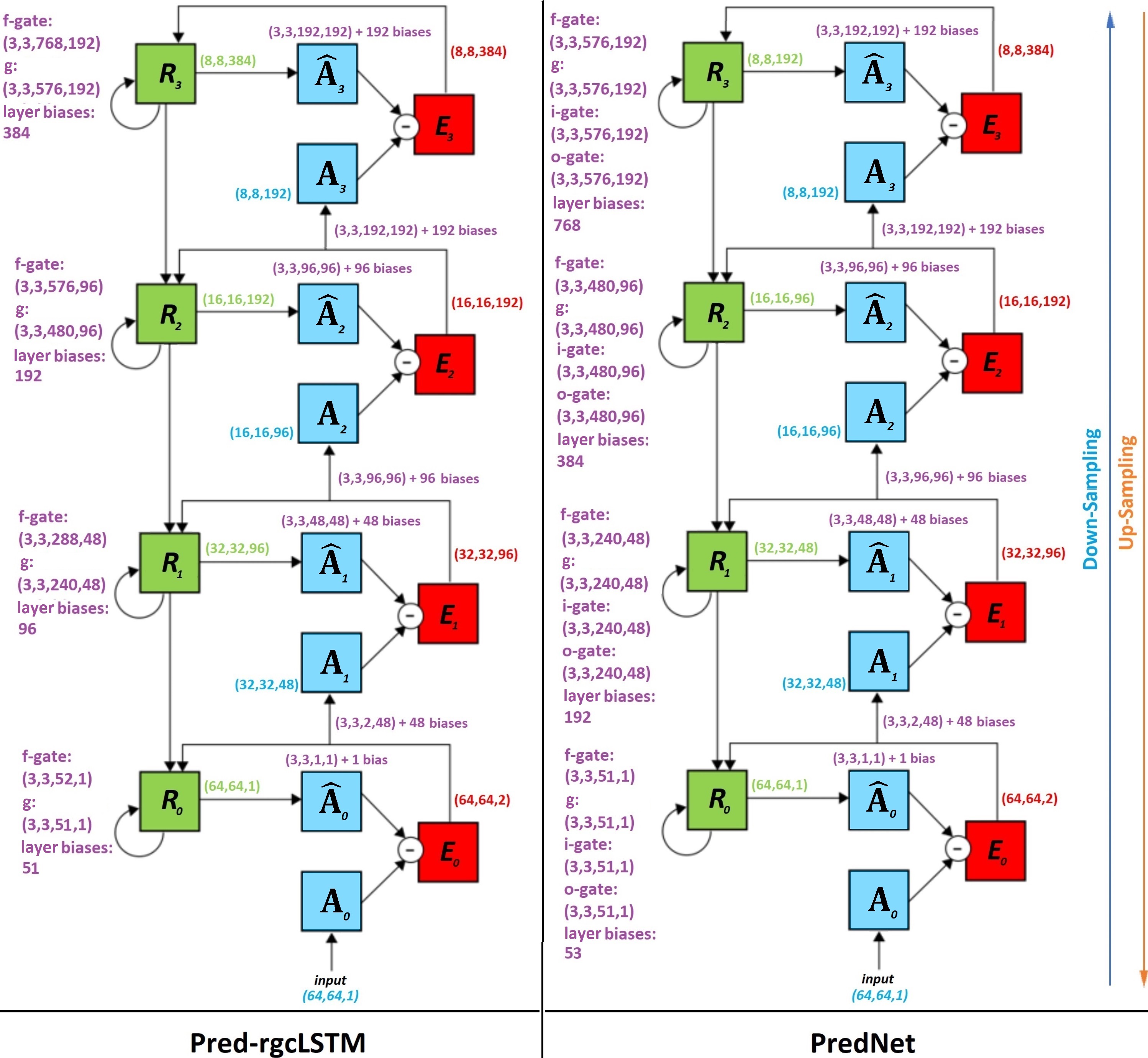}
	\caption{The weight counts of each 
	component of both Pred-rgcLSTM and 
	PredNet
	are
	based on
	reference~\citep{Lotter2017}.
	The input (Moving MNIST) is a $64\times 64$ gray scale image with one channel.
	A tuple for a gate in the R module of the form $(x,x,y,z)$ signifies a count of $z$ convolution kernels
	with dimension $x\times x$,
	each with $y$ input channels.
	}
	\label{rgcLSTM_weights_graph}
\end{figure*}

Although Figure~\ref{figOverview} shows the 
lowest
layer (because the input is a video frame),
we still include layer index subscripts, $l$, because the higher layers have 
the same
structure (see Figure~\ref{rgcLSTM_weights_graph})
although the channel counts and image dimensions differ for each layer.
The prediction error module drives the gated-recurrent learning
(via backpropagation by combining the prediction errors into a cost function).
The module calculates the error between the input frame $A_l$ and the predicted 
output frame $\widehat{A}_l$.
The 
convolution module converts the output of the rgcLSTM, $h_l^{(t)}$,
to $\widehat{A}_l^{(t)}$ so the dimensions are compatible with $A_l$ for pixelwise subtraction.

The prediction error module stacks the pixelwise difference 
(Eqn.~\ref{eqnErrorStack}) 
between the predicted image frame $\widehat{A}_l$ and the 
actual frame $A_l$ by subtracting 
the two frames and applying the $ReLU$ 
as follows:

\begin{align}
err1 &= \mathrm{ReLU}(\widehat{A}_l-A_l)\label{eqnError1}\\
err2 &= \mathrm{ReLU}(A_l-\widehat{A}_l)\label{eqnError2}\\
e_l &= [err1, err2]\label{eqnErrorStack}
\end{align}


\begin{figure*}
	\centering
	\includegraphics[width=13.cm,height=6cm]{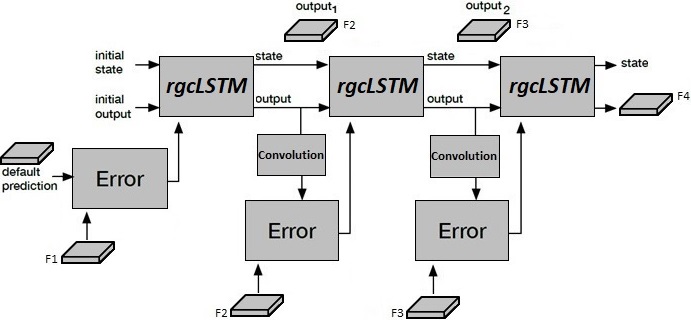}
	\caption{Three unrollment steps for the lowest layer of Pred-rgcLSTM using three input frames: F1, F2, and F3.
		F4 is the next-frame prediction.}
	\label{figUnrollments}
\end{figure*}

\noindent
In the case of gray scale,
$e_\mathrm{l}$ is a stack of the two error images 
encoding negative and positive prediction errors according to
Eqns.~\ref{eqnError1} and~\ref{eqnError2}.
The 
$e_l$
feedback is sent to the rgcLSTM, $R_l$, and to the next higher 
layer $R_{l+1}$ after applying 
downsampling 
to obtain image
dimensions 
that match
the next layer 
dimensions
(see Fig~\ref{rgcLSTM_weights_graph}).
Inputs to the rgcLSTM module are its internal recurrent input and its error feedback. 
In the multi-layer case, the input contains an additional parameter which is the feedback from the directly 
higher rgcLSTM block, $r_\mathrm{upper}$. 
These inputs 
are stacked
to fit into the rgcLSTM block.

The rgcLSTM block learns the spatiotemporal changes using the training input data to predict future frames. 
The update process is a function of the sequential input within one layer of the proposed model as shown in Figure~\ref{figUnrollments}. 
Figure~\ref{figUnrollments}
shows three unrollment (unfolding in time) steps and 
tracks
the model from 
the leftmost (earliest) part of 
the diagram. 
In the beginning, the error module evaluates the prediction error by the difference between the assumed default prediction and the first input which is the first video frame. 
This prediction error then goes to the rgcLSTM module as an input. 
The rgcLSTM module processes this input, its initial state,
and initial output through its internal gates to produce the next predicted video frame as its output. 
Next, the current output of the rgcLSTM module passes through the error module to evaluate the next prediction error which is forwarded to the next unfolded rgcLSTM module. 
At this time the prediction error acts as a correction factor for the rgcLSTM module which guides the rgcLSTM module to 
adjust its weights and state according to the current frame and the previous prediction of this frame. 
The process repeats 
until the 
final predicted frame is processed.

Since our model targets spatiotemporal prediction, we used \textit{hardSig} as the recurrent activation function 
and $\mathit{tanh}$ as the output activation function. 
\textit{hardSig} was calculated as:
\begin{equation}
\label{eqnHardsig}
\mathrm{hardSig}(x) = \max( \min(0.25\:x + 0.5,1), 0) .
\end{equation}

\noindent
We chose the $\mathit{hardSig}$ because it has shown better empirical results in our earlier experiments
than the sigmoid in LSTM spatiotemporal prediction models~\citep{elsayed2018empirical,Elsayed2019}. 
The hard saturation may help it escape from local minima~\citep{Gulcehre2016}.

\begin{table*}[t]%
\begin{center}
		\caption{Definition of terms.\label{term_definition_table}}
		\centering
		\begin{tabular*}{300pt}{@{\extracolsep\fill}llD{.}{.}{3}c@{\extracolsep\fill}}
			\toprule
			\textbf{Symbol} & \textbf{Meaning} \\
			\midrule
			$h_l$&Eqn.~\ref{rgcLSTM_output_eqn}.\\
			$e_l$&Eqn.~\ref{eqnErrorStack}. $e_0 = x$.\\
			$c_l$&Eqn.~\ref{memory_cell}.\\
			$r_{\mathrm{upper}(l)}$&feedback 
			
			to $R_l$ 
			
			from layer $l+1$.\\
			$R_l$&dimension of gate input stack 
			($\gamma + 2\kappa$).
			\\
			$\mathit{Ra}_l$&dimension activation input stack 
			($\gamma+\kappa$).
			\\
			$f_l$&Eqn.~\ref{forget_gate_eqn}.\\
			$g_l$&Eqn.~\ref{memoryActEqn}.\\
			$i_l$&PredNet input gate for layer $l$.\\
			$o_l$&PredNet output gate for layer $l$.\\
			$A_l$&next frame input for layer $l$.\\
			$\widehat{A}_l$&predicted next frame input for layer $l$.\\
			$\mathit{downsample}_l$&convolutional downsampling from layer $l$ to $l+1$.\\
			$\mathit{biases}_l$&total number of biases in layer $l$.\\
			\bottomrule
		\end{tabular*}
\end{center}
\end{table*}

We now explain the full multi-layer model shown in Figure~\ref{rgcLSTM_weights_graph}.
Table~\ref{term_definition_table} 
lists and
defines the terms used in the model.
The input dimensions of each layer of the four-layered Pred-rgcLSTM and PredNet models are shown in Table~\ref{input_summary_table}. 
$h_l$ is the recurrent output of the rgcLSTM or cLSTM, $e_l$ is the error feedback, $c_l$ is the memory cell of the gated unit (equivalent to $c_{(t-1)}$), 
and $r_\mathrm{upper(l)}$ is the higher-level feedback from layer $l+1$. 
$R_l$ is the total dimension of each the forget gate $f^{(t)}$input stack and $\mathit{Ra}_0$ is the total dimension 
of the input stack to the activation unit $g^{(t)}$. 
Table~\ref{summary_weights_table} and Figure~\ref{rgcLSTM_weights_graph} show the dimensions of the weights and biases 
in each layer 
of
rgcLSTM-PredNet as compared to the
original PredNet.
This is shown
for each layer index $l \in \{0,1,2,3\}$.~$f_l$, $i_l$, $o_l$ are the forget, 
input and output gates, respectively. $\mathit{g}_l$ is the activation unit. 
$\mathit{downsample}_l$ is the transition kernel from the lower layer to the higher layer. $\widehat{A}_l$ is the convolution kernel applied to the rgcLSTM output or the cLSTM output. 
$\mathit{Biases}$ are the total number of biases in each layer (i.e., including the gate, activation, 
downsampling, and the transition between the output of the rgcLSTM and the error module).

In Table~\ref{input_summary_table}, Table~\ref{summary_weights_table}, 
and Figure~\ref{rgcLSTM_weights_graph} we assume that the model input is a gray-scale image of size $64\times64$ pixels and has a one input channel depth.
For the purpose of counting 
the
trainable parameters, the values of $m$, $\gamma$, $\kappa$, and $n$ 
from Section~\ref{sec:rgcLSTM} are given 
in Table~\ref{tabParamCounts} for
all four layers in the model. 

\begin{table*}[t]%
\begin{center}
		\caption{Dimensions of input components to Pred-rgcLSTM and PredNet for Moving MNIST.
		A tuple of the form $(x,x,y)$ signifies an $x\times x$ kernel with $y$ input channels.
		Both models have four layers and the lowest is indexed by the subscript zero.
		\label{input_summary_table}}
		\centering
		\begin{tabular*}{300pt}{@{\extracolsep\fill}lllD{.}{.}{3}c@{\extracolsep\fill}}
			\toprule
				&\multicolumn{2}{@{}c@{}}{\textbf{Dimensions}} \\ 
				\cmidrule{2-3}
			\textbf{Parameter} & \textbf{Pred-rgcLSTM} & \textbf{PredNet}\\
			\midrule
			$h_0$&(64, 64, 1)&(64, 64, 1)\\
			$e_0$&(64, 64, 2)&(64, 64, 2)\\
			$c_0$&(64, 64, 1)&N/A\\
			$r_\mathrm{upper(0)}$&(64, 64, 48)&(64, 64, 48)\\
			$R_0$ gate input stack&(64, 64, 52)&(64, 64, 51)\\	
			$Ra_0$ activation input stack&(64, 64, 51)&(64, 64, 51)\\	
			\hline
			$h_1$&(32, 32, 48)&(32, 32, 48)\\
			$e_1$&(32, 32, 96)&(32, 32, 96)\\
			$c_1$&(32, 32, 48)&N/A\\
			$r_\mathrm{upper(1)}$&(32, 32, 96)&(64, 64, 96)\\
			$R_1$ input stack&(32, 32, 288)&(32, 32, 240)\\	
			$Ra_1$ activation input stack&(32, 32, 240)&(32, 32, 240)\\	
			\hline	
			$h_2$&(16, 16, 96)&(16, 16, 96)\\
			$e_2$&(16, 16, 192)&(16, 16, 192)\\
			$c_2$&(16, 16, 96)&N/A\\
			$r_\mathrm{upper(2)}$&(16, 16, 192)&(16, 16, 192)\\
			$R_2$ input stack&(16, 16, 576)&(16, 16, 480)\\	
			$Ra_2$ activation input stack&(16, 16, 480)&(16, 16, 480)\\
			\hline	
			$h_3$&(8, 8, 192)&(8, 8, 192)\\
			$e_3$&(8, 8, 384)&(8, 8, 384)\\
			$c_3$&(8, 8, 192)&N/A\\
			$R_3$ input stack&(8, 8, 768)&(8, 8, 576)\\	
			$Ra_3$ activation input stack&(8, 8, 576)&(8, 8, 576)\\	
			\bottomrule
		\end{tabular*}
\end{center}
\end{table*}

\begin{center}
	\begin{table*}[t]%
		\caption{Dimensions of kernel (weight) components of Pred-rgcLSTM and PredNet for Moving MNIST.\label{summary_weights_table}}
		\centering
		\begin{tabular*}{300pt}{@{\extracolsep\fill}lllD{.}{.}{3}c@{\extracolsep\fill}}
			\toprule
			&\multicolumn{2}{@{}c@{}}{\textbf{Dimensions}} \\ 
			\cmidrule{2-3}
			\textbf{Kernel} & \textbf{Pred-rgcLSTM} & \textbf{PredNet}\\
			\midrule
			$f_0$&(3, 3, 52, 1)&(3, 3, 51, 1)\\
			$g_0$&(3, 3, 51, 1)&(3, 3, 51, 1)\\
			$i_0$&N/A&(3, 3, 51, 1)\\
			$o_0$&N/A&(3, 3, 51, 1)\\
			$\widehat{A}_0$&(3, 3, 1, 1)&(3, 3, 1, 1)\\	
			$downsample_0$&(3, 3, 2, 48)&(3, 3, 2, 48)\\
			$biases_0$&51&53\\
			\hline
			$f_1$&(3, 3, 288, 48)&(3, 3, 240, 48)\\
			$g_1$&(3, 3, 240, 48)&(3, 3, 240, 48)\\
			$i_1$&N/A&(3, 3, 240, 48)\\
			$o_1$&N/A&(3, 3, 240, 48)\\
			$\widehat{A}_1$&(3, 3, 48, 48)&(3, 3, 48, 48)\\	
			$downsample_1$&(3, 3, 96, 96)&(3, 3, 96, 96)\\
			$biases_1$&240&336\\
			\hline	
			$f_2$&(3, 3, 576, 96)&(3, 3, 480, 96)\\
			$g_2$&(3, 3, 480, 96)&(3, 3, 480, 96)\\
			$i_2$&N/A&(3, 3, 480, 96)\\
			$o_2$&N/A&(3, 3, 480, 96)\\
			$\widehat{A}_2$&(3, 3, 96, 96)&(3, 3, 96, 96)\\	
			$downsample_2$&(3, 3, 192, 192)&(3, 3, 192, 192)\\
			$biases_2$&480&672\\	
			\hline	
			$f_3$&(3, 3, 768, 192)&(3, 3, 576, 192)\\
			$g_3$&(3, 3, 576, 192)&(3, 3, 576, 192)\\
			$i_3$&N/A&(3, 3, 576, 192)\\
			$o_3$&N/A&(3, 3, 576, 192)\\
			$\widehat{A}_3$&(3, 3, 192, 192)&(3, 3, 192, 192)\\	
			$biases_3$&576&960\\
			\hline	
			Total training params&\textbf{4,316,235}&6,909,834\\
			\bottomrule
		\end{tabular*}
	\end{table*}
\end{center}

\begin{center}
	\begin{table*}[t]%
		\caption{Dimension parameters for the gated module (either rgcLSTM or cLSTM) for each of the four layers for 
		the
		moving MNIST
		dataset.\label{tabParamCounts}
		}
		\centering
		\begin{tabular*}{300pt}{@{\extracolsep\fill}r|rrrrD{.}{.}{3}c@{\extracolsep\fill}}
			\toprule
			\textbf{Parameter} & \textbf{Layer$_0$}& \textbf{Layer$_1$} & \textbf{Layer$_2$}& \textbf{Layer$_3$}\\
			\midrule
			$m$       & 3&  3&  3&  3\\
			$\gamma$  &50&192&384&384\\
			$\kappa$  & 1& 96&192&384\\
			$n$       & 1& 96&192&384\\
			\bottomrule
		\end{tabular*}
	\end{table*}
\end{center}

\begin{center}
	\begin{table*}[t]%
		\caption{Experimental conditions in the Moving MNIST experiment for recurrent gate-based blocks. 
		These correspond to experiments shown in Table~\ref{mnist_comparision_1}.
		    N/A means there cannot be peephole connections
			because there is no memory cell state.\label{units_compare}}
		\centering
		\begin{tabular*}{300pt}{@{\extracolsep\fill}lcccD{.}{.}{3}c@{\extracolsep\fill}}
			\toprule
			\textbf{Model} & \textbf{Memory Cell}& \textbf{Gates} & \textbf{Peephole}\\
			\midrule
			rgcLSTM& Yes&1&Yes\\
			LSTM& Yes&3&No\\
			PLSTM& Yes&3&Yes\\
			GRU& No&2&N/A\\
			MGU& No&1&N/A\\
			\bottomrule
		\end{tabular*}
	\end{table*}
\end{center}

\subsection{Keeping the Gate Output within a Functional Operating Range}
From Figure~\ref{rgcLSTM_architecture} we see that if the module gate value drops to zero,
the rgcLSTM cell does an extreme reset:
the memory state is forgotten, the current input is ignored, and the output is zero.
Within
the predictive coding architecture, 
there are several factors that help
keep the gate value within a functional range to avoid this.
First, the input image pixel values are normalized to the range $\left[0,1\right]$ and these
become the 
target
values for $E_0$ in the lowest layer of the network (Eqns.~\ref{eqnError1} and \ref{eqnError2})\@.
The values for
$\widehat{A}_l$ are also kept in the range $\left[0,1\right]$.
The $\mathrm{ReLU}$
serves as a rectifier to keep
the $e_l$ within the range $[0,1]$.
These form the inputs $x=e_l$ to the rgcLSTM\@.
Within the rgcLSTM, the values of $h$ and $c$ (Figure~\ref{rgcLSTM_architecture}) are constrained
to fall in the range $(-1, 1)$ because they are outputs of a $\tanh$ or a
gate
modulated $\tanh$.
Finally, the module gate uses a hard sigmoid configured to be linear in the range $[-2,2]$.
The inputs to the hard sigmoid stay well within this range, thereby keeping the gate functional.

\section{Methods}
\label{sec_methodsForExp1AndExp2}
 
Experiments were performed to compare our rgcLSTM used in Pred-rgcLSTM with the standard cLSTM used in PredNet. 
We show that our rgcLSTM achieves the same or better accuracy than the standard cLSTM using a smaller 
parameter budget, less training time, and smaller memory requirements.
To build Pred-rgcLSTM, we modified the code for PredNet which is available at
\texttt{https://github.com/coxlab/prednet}. 
In all of our simulations and experiments, the gate activation for the rgcLSTM cells
was hardsig set to be linear over the range $\left[-2, 2\right]$.
To make the timing results comparable for both models, 
the only modification to the code was to replace the cLSTM with the rgcLSTM of compatible 
size to ensure the comparisons were fair. Our rgcLSTM source code is available at the following link \texttt{https://github.com/NellyElsayed/rgcLSTM}.

Two experiments
were conducted
on
next-frame(s) prediction of a spatiotemporal (video) sequence: 
one for a gray-scale dataset (moving MNIST) and the other for an RGB dataset (KITTI traffic).
A third experiment is described later.
The moving MNIST dataset~\citep{srivastava2015unsupervised} is a good example of how the model can determine movement of two specific objects within 
a frame and how the model can handle new object shapes and movement directions.
The KITTI dataset~\citep{geiger2013vision} requires the model to 
track
several different moving and non-moving objects within a frame
The 
roof-mounted camera records vehicle traffic which has 
both
static objects (scene) and dynamic objects (moving vehicles). 
Also, it shows how the model can deal with 
3--channel
RGB videos. For both experiments, we trained our model using four layers like that shown in Figure~\ref{figOverview}. 
The number of parameters and inputs are the same except for the first layer in each experiment, due to the input different 
input sizes for moving MNIST versus KITTI\@.
The loss function was a weighted combination of the mean-squared-error output by the modules $E_0 \ldots E_3$.
Lotter et al.~\citep{Lotter2017} found that the best weighting was $\left[1, 0, 0, 0\right]$, so our experiments used that choice.
Both experiments
used the Adam optimizer~\citep{kingma2014adam} with an initial learning rate $\alpha= 0.001$ and a decay factor of $10$ after half of the training process, and $\beta_1 = 0.9$ and $\beta_1 = 0.999$. 
Frame size was downsampled by factor of $2$ moving upwards through the layers and 
upsampled by factor of 2 moving down the layers. 

For training elapsed time comparisons in Experiment~1, 
the model was trained on an Intel(R) Core i7-7820HK CPU with $32$GB memory and an NVIDIA GeForce GTX 1080 graphics card. 
For Experiment~2, both Pred-rgcLSTM and PredNet were trained on Intel(R) Core i7-6700 @4.00GHx8 processor with 
$64$ GB memory and NVIDIA GeForce 
GTX
980 Ti/PCle/SSE2 graphics card.

The methods for Experiment 3 are different than Experiments~1 and~2, so they are described later in Section~\ref{sec_methodForExp3}.

\subsection{Experiment 1: The Moving MNIST Dataset}
\label{sec_Exp1_MovingMNIST}

Although the original 
PredNet 
model was not tested on the Moving MNIST dataset,
we considered it appropriate to start with this simpler dataset for our experiments.
This experiment compared the video prediction performance of our rgcLSTM-PredNet with the 
original PredNet.

\subsubsection{Method}

The Moving MNIST dataset consists of $10,000$ video sequences of two randomly selected moving digits sampled from the 
original
MNIST dataset. 
Each sequence is $20$ frames long, with
a frame size of $64$x$64$ 
grayscale
pixels. 
Thus,
each frame has a depth of one channel. 
We divided the dataset into $6,000$ video sequences for training,~$3,000$ for validation and $2,000$ for testing.
The training process was completed in one epoch. 
The number of core-channels in each layer were $1$, $48$, $96$, and $192$ respectively. 
Hence, the dimensions of weights and inputs are shown in Tables~\ref{input_summary_table} and~\ref{summary_weights_table}, and Figure~\ref{rgcLSTM_weights_graph}.

Our simulations are performed using rgcLSTM-PredNet and 
PredNet
but, in the results section,
we compare our simulation results
with several
spatiotemporal prediction approaches that use recurrent gated units.
Table~\ref{units_compare} shows the properties of the models we consider.
These properties are the existence of a memory cell, the number of gates, and
the existence of peephole connections.
The models we 
compare
are our novel
rgcLSTM, 
the standard LSTM~\citep{gers2000learning}, the peephole LSTM (PLSTM)~\citep{gers2002learning}, the gated recurrent unit (GRU)~\citep{cho2014properties}, and the minimal gated unit (MGU) \citep{zhou2016minimal}.

\begin{center}
	\begin{table*}[t]%
		\caption{Moving MNIST performance comparison.\label{mnist_comparision_1}}
		\centering
		\begin{tabular*}{300pt}{@{\extracolsep\fill}lrcccD{.}{.}{3}c@{\extracolsep\fill}}
			\toprule
			\textbf{Model} & \textbf{Type}& \textbf{MSE}& \textbf{MAE}& \textbf{SSIM}\\
			\midrule
			FCLSTM~\citep{Shi2015,srivastava2015unsupervised}&LSTM& 1.865& 2.094& 0.690\\
			CDNA~\citep{wang2018predrnn++,finn2016unsupervised}&cLSTM& 0.974& 1.753& 0.721\\
			DFN~\citep{wang2018predrnn++,jia2016dynamic}&DynamicFilters& 0.890& 1.728& 0.726\\
			VPN~\citep{wang2018predrnn++,kalchbrenner2016video}&cLSTM& 0.641& 1.310& 0.870\\
			ConvLSTM~\citep{wang2018predrnn++,Shi2015}&convLSTM& 1.420& 1.829&0.707\\
			ConvGRU~\citep{wang2018predrnn++,shi2017deep}&convGRU&1.254& 2.254& 0.601\\
			TrajGRU~\citep{wang2018predrnn++,shi2017deep}&convGRU& 1.138& 1.901& 0.713\\
			PredRNN++~\citep{wang2018predrnn++}&convLSTM& 0.465& 1.068& 0.898\\
			PredRNN~\citep{wang2017predrnn,wang2018predrnn++}&convLSTM& 0.568& 1.261& 0.867\\
			PredNet~\citep{Lotter2017}&cLSTM& 0.011& 0.049&0.915\\
			Pred-rgcLSTM&rgcLSTM& \textbf{0.009}& \textbf{0.017}& \textbf{0.924}\\
			\bottomrule
		\end{tabular*}
		
	\end{table*}
\end{center}

\begin{center}
	\begin{table*}[t]%
		\caption{Moving MNIST average elapsed training time and standard error (n=15).\label{mnist_comparision_3}}
		\centering
		\begin{tabular*}{300pt}{@{\extracolsep\fill}lccD{.}{.}{3}c@{\extracolsep\fill}}
			\toprule
			\textbf{Model} & \textbf{Training Time (m)}& \textbf{Standard Error (SE)}\\
			\midrule
			PredNet&100.328&0.586532\\
		
			Pred-rgcLSTM& \textbf{73.561}&\textbf{0.297411}\\ 
			\bottomrule
		\end{tabular*}
	\end{table*}
\end{center}

\subsubsection{Results of Experiment 1}
Table~\ref{mnist_comparision_1} shows
performance comparisons among 
several
models
for moving MNIST.
We compared Pred-rgcLSTM with several other unsupervised recurrent gate-based models 
(i.e., either LSTM or GRU based models) for the moving MNIST dataset.
Performance measures included mean squared error (MSE), mean absolute error (MAE),
and the structural similarity index (SSIM)~\citep{ssim_reference} which is a measure for image quality similarity structure 
between predicted and actual images. 
We also compared our model to PredNet.
For the remaining models that we were unable to test due to hardware limitations 
we obtained the results from their published work. 
All 
of
the models are convolutional variants. 
Our Pred-rgcLSTM shows a reduction of both the MSE and MAE compared to all of the other models tested.
This includes
FCLSTM~\citep{srivastava2015unsupervised}, CDNA~\citep{finn2016unsupervised}, DFN~\citep{jia2016dynamic}, 
VPN~\citep{kalchbrenner2016video}, ConvLSTM~\citep{Shi2015}, ConvGRU, TrajGRU~\citep{shi2017deep}, PredRNN~\citep{wang2017predrnn}, 
PredRNN++~\citep{wang2018predrnn++} and 
PredNet~\citep{Lotter2017}.
The main comparison is between our rgcLSTM block and the cLSTM block using the same model architecture. 
Our model also has the best structural similarity index measurement (SSIM) among the tested models.

\begin{center}
	\begin{table*}[t]%
		\caption{Moving MNIST memory required to save Pred-rgcLSTM and PredNet trained parameters.\label{mnist_memory}}
		\centering
		\begin{tabular*}{300pt}{@{\extracolsep\fill}lcD{.}{.}{3}c@{\extracolsep\fill}}
			\toprule
			\textbf{Model} & \textbf{Memory}\\
			\midrule
			PredNet& 81.068 MB\\
			Pred-rgcLSTM& \textbf{50.650} MB\\
			\bottomrule
		\end{tabular*}
		
	\end{table*}
\end{center}

The training time in minutes for the Pred-rgcLSTM and PredNet models for one batch of $6,000$ trials
is shown
in Table~\ref{mnist_comparision_3}.
Replacing
the rgcLSTM 
cell
with
the cLSTM reduced elapsed training time by about $26\%$ for 
training moving MNIST for one epoch. 
Table~\ref{mnist_comparision_3} also shows the SE of model training time for both the cLSTM and our rgcLSTM model. 
The sample
size
was $n=15$ for each model. 

For the 
cLSTM, 
the number of parameters is calculated by:
\begin{align}
\begin{split}
\mathrm{cLSTM}_\mathrm{cell}^\#& = f_\mathrm{gate}^\# + g_\mathrm{update}^\# + i_\mathrm{gate}^\# + o_\mathrm{update}^\#\\
& = 4 \cdot (\left(m^2(\gamma+\kappa)+1\right)\cdot n)\\
\end{split}
\label{eqn_cLSTMcount}
\end{align}

\noindent
Here,
$i_\mathrm{gate}^\#$ and $o_\mathrm{gate}^\#$ are the number of trainable parameters for the input and output gates of the standard convLSTM cell. 
The multiplication by four is due to the input update activation, forget gate, input gate, 
and output gate that each has the same number of trainable parameters. 
Each cLSTM gate uses the same number of parameters as the $g_\mathrm{update}$ gate in the rgcLSTM.
Recall that Eqn.~\ref{parameters_count} gives the parameter count for our rgcLSTM\@.
By comparing Eqns.~\ref{parameters_count} and~\ref{eqn_cLSTMcount}, we see that
the number of trainable parameters is reduced approximately by $40\%$
in the rgcLSTM compared to the cLSTM\@.

We also compared the memory 
needed
(MB) to save the Pred-rgcLSTM and 
PredNet trained models.
This is shown in Table~\ref{mnist_memory}.
The training parameters of the Pred-rgcLSTM and the 
PredNet
were saved using the Keras API and the size of each file was examined. 
Pred-rgcLSTM required significantly less memory than PredNet.

Using the Keras API \texttt{model.count\_params()}~\citep{kerasAPI},
we 
compared the number of training parameters used in Pred-rgcLSTM and PredNet of the implemented models in Table~\ref{mnist_comparision_2}. 
For the remaining models in Table~\ref{mnist_comparision_2}, we obtained the results from their published work~\citep{shi2017deep,Shi2015,srivastava2015unsupervised} which also were counted by using the same Keras API method~\citep{kerasAPI}.
The empirical parameter counts for the implemented PredNet and Pred-rgcModels match the counts
given in the design specifications shown in Table~\ref{summary_weights_table}, giving some measure of confidence that 
these two
models  are implemented correctly.
Our model uses the fewest parameters of the tested models. 
Our model has fewer trainable parameters approximately by $40\%$-$50\%$ than the ConvGRU and ConvLSTM respectively. 
Our Pred-rgcLSTM model that is based on rgcLSTM block has approximately $40\%$ fewer trainable parameters than the 
original PredNet model that uses cLSTM modules.
This latter claim can be confirmed by examining the calculations in Table~\ref{summary_weights_table}.

\begin{center}
	\begin{table*}[t]%
		\caption{Moving MNIST: the number of trainable parameters in each model.\label{mnist_comparision_2}}
		\centering
		\begin{tabular*}{300pt}{@{\extracolsep\fill}lrD{.}{.}{3}c@{\extracolsep\fill}}
			\toprule
			\textbf{Model} & \textbf{Trainable Parameters}\\
			\midrule
			PredNet& 6,909,834\\
			ConvGRU& 8,010,000\\
			TrajGRU& 4,770,000\\
			ConvLSTM& 7,585,296\\
			FCLSTM &142,667,776\\
			Pred-rgcLSTM& \textbf{4,316,235}\\
			\bottomrule
		\end{tabular*}
		
	\end{table*}
\end{center}

\begin{figure*}
	\centering
	\includegraphics[width=17cm,height=8cm]{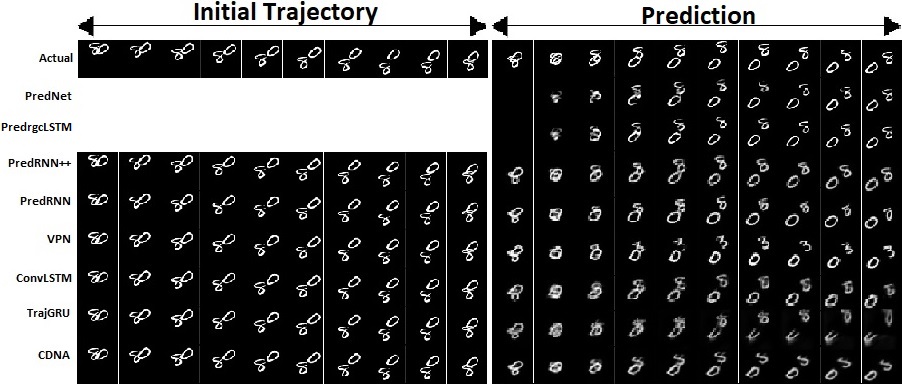}
	\caption{Visual results of Moving MNIST predictions after training based on our rgcLSTM and other models.}
	\label{mnist_visual_results}
\end{figure*}

\begin{figure}
	\centering
	\includegraphics[width=13cm,height=5cm]{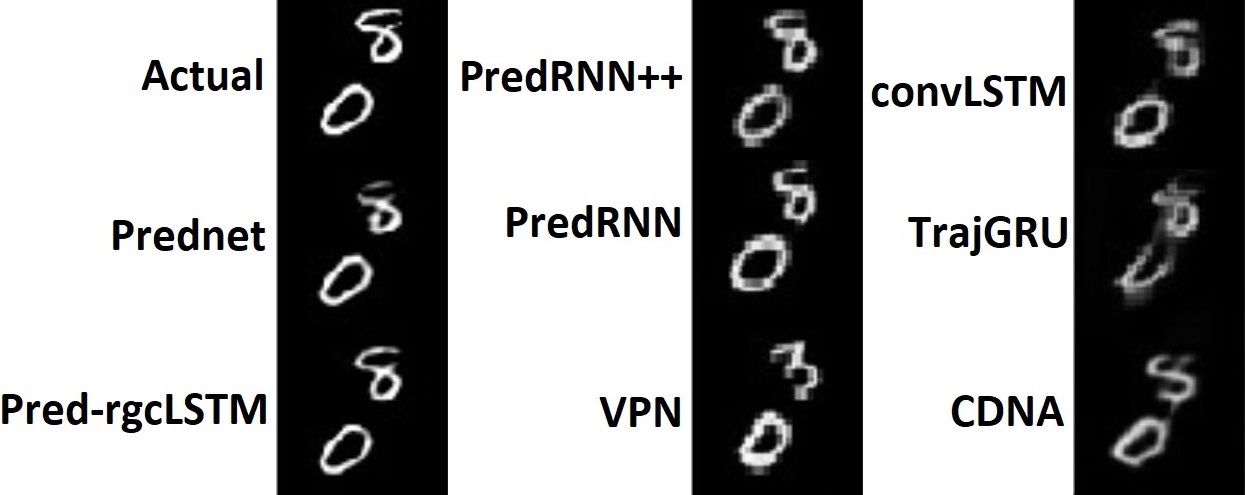}
	\caption{Magnified visual result of Moving MNIST predictions for prediction time step 6 after training based on our rgcLSTM and other models.
}
	\label{zoom_mnist_visual_results}
\end{figure}

For Experiment~1,
an example prediction
for the moving MNIST dataset is shown in Figure~\ref{mnist_visual_results} for nine different models. 
In contrast to most of the models, neither PredNet nor our Pred-rgcLSTM require an initial input trajectory for predicting the  upcoming frames.
They use only the current frame and current state of the trained LSTM to predict the next frame.
Our rgcLSTM demonstrates comparable results with the other models and finds better visual contours than the cLSTM used in PredNet. 
To see the improved prediction in better detail,
Figure~\ref{zoom_mnist_visual_results} magnifies the predicted images (taken from Figure~\ref{mnist_visual_results})
for prediction time step six for each model.

\begin{center}
	\begin{table*}[t]%
		\caption{KITTI dataset experiment dimensions of kernel (weight) components of the first layer of Pred-rgcLSTM and PredNet.\label{kitti_weights}}
		\centering
		\begin{tabular*}{300pt}{@{\extracolsep\fill}lllD{.}{.}{3}c@{\extracolsep\fill}}
			\toprule
			\textbf{Kernel} & \textbf{Pred-rgcLSTM} & \textbf{PredNet}\\
			\midrule
			$f_0$&(3, 3, 60, 3)&(3, 3, 57, 3)\\
			$g_0$&(3, 3, 57, 3)&(3, 3, 57, 3)\\
			$i_0$&N/A&(3, 3, 57, 3)\\
			$o_0$&N/A&(3, 3, 57, 3)\\
			$\widehat{A}_0$&(3, 3, 3, 3)&(3, 3, 3, 3)\\	
			$\mathit{downsample}_0$&(3, 3, 6, 48)&(3, 3, 6, 48)\\
			$\mathit{biases}_0$&60&57\\
			\bottomrule
		\end{tabular*}
		
	\end{table*}
\end{center}

\begin{center}
	\begin{table*}[t]%
		\caption{KITTI dataset experiment dimensions of input component of Pred-rgcLSTM and PredNet.
		The N/A entries indicate that PredNet does not have peephole connections.
		\label{kitti_inputs}}
		\centering
		\begin{tabular*}{300pt}{@{\extracolsep\fill}lllD{.}{.}{3}c@{\extracolsep\fill}}
			\toprule
			\textbf{Parameter} & \textbf{Pred-rgcLSTM} & \textbf{PredNet}\\
			\midrule
			$h_0$&(128, 160, 3)&(128, 160, 3)\\
			$e_0$&(128, 160, 6)&(128, 160, 6)\\
			$c_0$&(128, 160, 3)&N/A\\
			$r_{upper(0)}$&(128, 160, 48)&(128, 160, 48)\\
			$R_0$ gate input stack&(128, 160, 60)&(128, 160, 57)\\	
			$Ra_0$ activation input stack&(128, 160, 57)&(128, 160, 57)\\	
			\hline
			$h_1$&(64, 80, 48)&(64, 80, 48)\\
			$e_1$&(64, 80, 96)&(64, 80, 96)\\
			$c_1$&(64, 80, 48)&N/A\\
			$r_{upper(1)}$&(64, 80, 96)&(64, 80, 96)\\
			$R_1$ input stack&(64, 80, 288)&(64, 80, 240)\\	
			$Ra_1$ activation input stack&(64, 80, 240)&(64, 80, 240)\\	
			\hline	
			$h_2$&(32, 40, 96)&(32, 40, 96)\\
			$e_2$&(32, 40, 192)&(32, 40, 192)\\
			$c_2$&(32, 40, 96)&N/A\\
			$r_{upper(2)}$&(32, 40, 192)&(32, 40, 192)\\
			$R_2$ input stack&(32, 40, 576)&(32, 40, 480)\\	
			$Ra_2$ activation input stack&(32, 40, 480)&(32, 40, 480)\\
			\hline	
			$h_3$&(16, 20, 192)&(16, 20, 192)\\
			$e_3$&(16, 20, 384)&(16, 20, 384)\\
			$c_3$&(16, 20, 192)&N/A\\
			$R_3$ input stack&(16, 20, 768)&(16, 20, 576)\\	
			$Ra_3$ activation input stack&(16, 20, 576)&(16, 20, 576)\\	
			\bottomrule
		\end{tabular*}
		
	\end{table*}
\end{center}

\subsubsection{Discussion of Experiment 1}
This experiment shows that the version of PredNet which uses our novel rgcLSTM
modules gives state-of-the-art video prediction performance in comparison to a range
of models, as seen in Table~\ref{mnist_comparision_1}.
In particular, its performance is on par with the original PredNet, which uses the cLSTM
instead of our rgcLSTM\@.

Furthermore, our rgcLSTM version of PredNet is significantly more cost effective than the original PredNet.
Specifically, the parameter count is reduced by about 40\%, the memory needed to save the trained parameters
is reduced by about 38\%, and the elapsed training time is reduced by about 27\%.

\subsection{Experiment 2: KITTI Dataset}
Experiment 2 was conducted to compare the rgcLSTM-PredNet performance with the original PredNet on
another dataset besides Moving MNIST\@.
The KITTI traffic dataset was chosen because it was used in the original PredNet experiments.
Its dimensions were resized as described at: http://github.com/coxlab/prednet.

\subsubsection{Method}
The experimental method is essentially the same as was used for the Moving MNIST dataset as described
at the beginning of Section~\ref{sec_methodsForExp1AndExp2}.
Any differences in method stem from modifying the architecture to accomodate the larger input dimensions
of the KITTI dataset.
The KITTI dataset~\citep{geiger2013vision}
is a collection of real life traffic videos captured by a roof-mounted camera on a car
driving in 
an urban environment. 
The data is divided into three categories: city, residential, and road. 
Each category contains $61$ different recorded sessions (i.e., different video sequences) which 
were divided into $57$ sessions for training and four for validation.
Simulations were performed for both our rgcLSTM-PredNet and the original cLSTM-PredNet.
For training, both 
models
used the same initialization, training, validation, and 
data down-sampling which was used by Lotter et al.~\citep{Lotter2017}. 
Each 
input image
frame size was $128\times160$ pixels,
where the image height is 128 pixels and the image width is 160,
and three RGB channels.
The total length of the training dataset (of the $57$ sessions) was approximately $41$K frames.
The model was trained for 150 epochs.

The architecture is very similar to that used in Experiment~1.
The number of kernels 
(output channels)
in each $R$ layer, indicated by the value of 
the last component of
$h_l$ in Table~\ref{kitti_inputs}, were $3$, $48$, $96$, and $192$, respectively. 
The resulting dimension changes for the weights in layer $R_0$ are shown in Table~\ref{kitti_weights}.
The other layer weight dimensions are unchanged from Table~\ref{summary_weights_table}. 
The input dimensions for each layer are changed in layer $R_0$ 
(because of the increased image channels and dimensions as compared to Moving MNIST)
in the width and height and number of input channels of the original video frame. 
However, the higher layers have changes only in the width and height 
dimensions.

\begin{center}
	\begin{table*}[t]%
		\caption{Performance comparison on the KITTI traffic video dataset.\label{kitti_table1}}
		\centering
		\begin{tabular*}{400pt}{@{\extracolsep\fill}lccccccD{.}{.}{3}c@{\extracolsep\fill}}
			\toprule
			\textbf{Model} & \textbf{Train Time} & \textbf{\# Params}& \textbf{MSE}& \textbf{MAE}& \textbf{SSIM}& \textbf{Memory}\\
			\midrule
			PredNet&230.982405&6,915,948&0.0044&0.0347&0.911&81.142 MB\\
			Pred-rgcLSTM& \textbf{140.530364}&\textbf{4,320,273}&\textbf{0.0035}&\textbf{0.030}&\textbf{0.932}&\textbf{50.697} MB\\ 
			\bottomrule
		\end{tabular*}
		
	\end{table*}
\end{center}

\begin{figure*}[t]
	\centering
	\includegraphics[keepaspectratio=true,width=17cm,height=9cm]{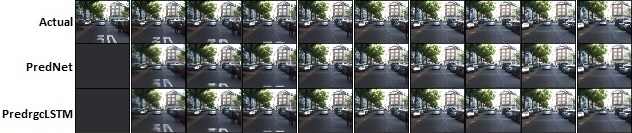}
	\caption{Next-frame prediction on the KITTI dataset. The predicted image is used as input for predicting the next frame.}
	\label{KITTI_results}
\end{figure*}
\begin{figure}[t]
	\centering
	\includegraphics[width=14cm,height=6cm]{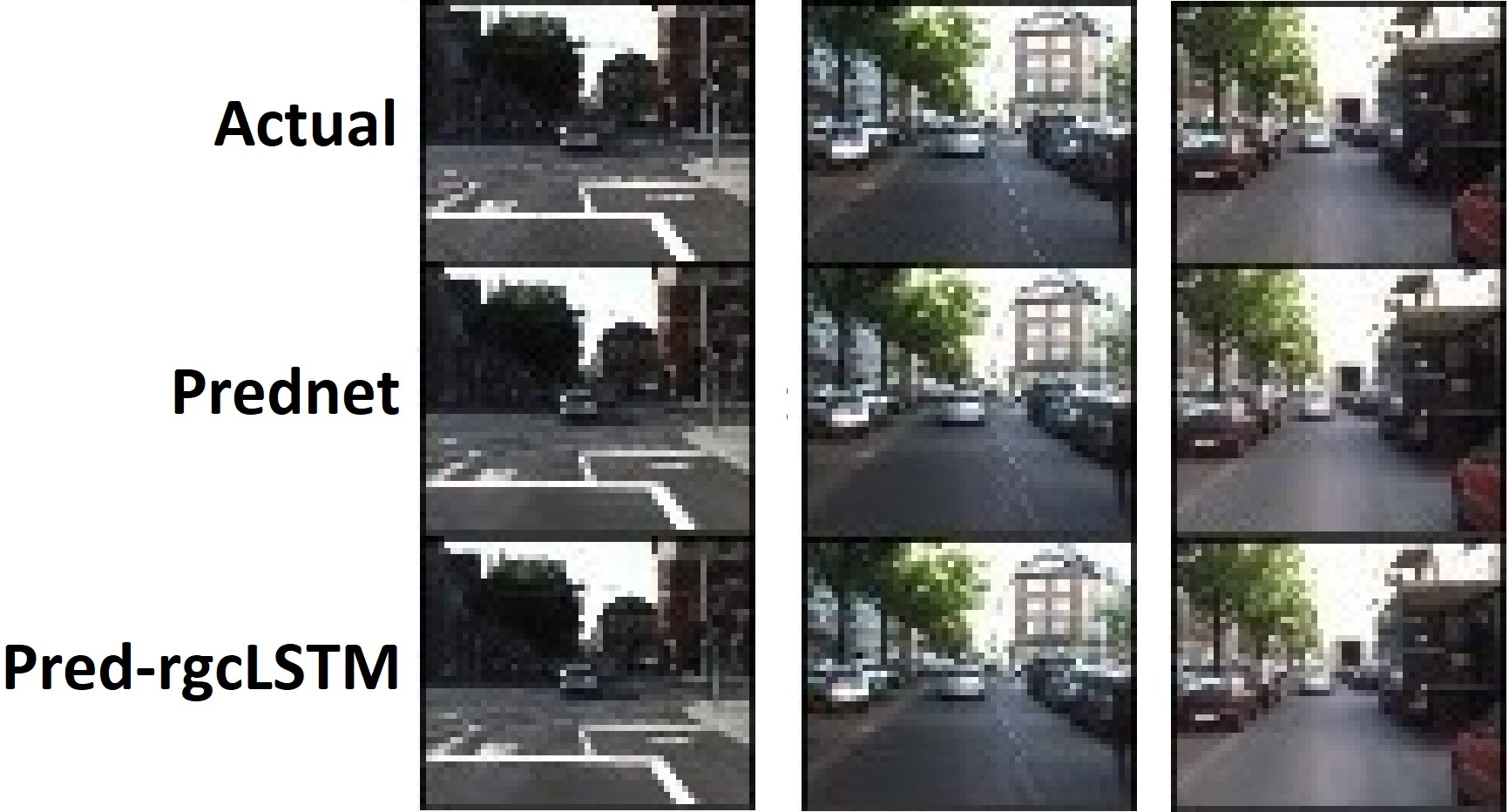}
	\caption{The magnified visual results for 
	the KITTI dataset prediction for the 6th prediction step for Pred-rgcLSTM and 
	PredNet.
	}
	\label{KITTI_zoom}
\end{figure}
\begin{figure*}
	\centering
	\includegraphics[keepaspectratio=true,width=18cm,height=5cm]{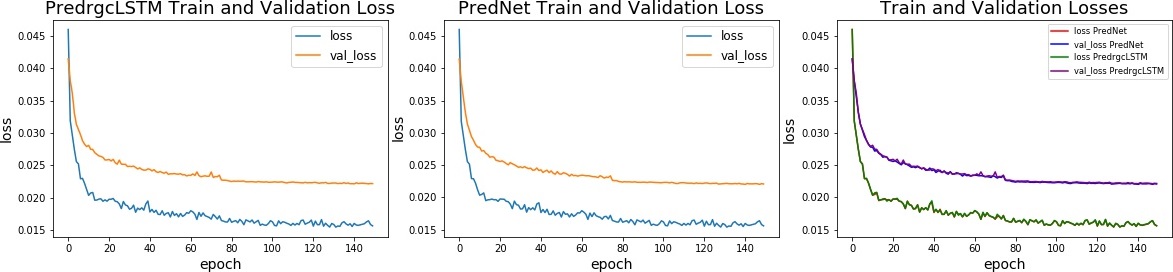}
	\caption{Example KITTI dataset training and validation losses for both Pred-rgcLSTM and PredNet.
		Left: Pred-rgcLSTM. Middle: 
		PredNet. 
		Right: both graphs are superimposed.
		}
	\label{kitti_train_loss}
\end{figure*}

\subsubsection{Results of Experiment 2}

Table~\ref{kitti_table1} shows
the results for the KITTI dataset benchmark including elapsed training time (minutes), number of trainable parameters, 
mean squared error (MSE), mean absolute error (MAE), 
and mean structural similarity index measurement (SSIM)~\citep{ssim_reference} of our rgcLSTM and cLSTM.
The sample size is $n=16$ for each of these measures. 
The rgcLSTM has a higher SSIM 
than
the cLSTM. 
Table~\ref{kitti_table1} also shows the memory 
needed (MB) to save the trainable model for both Pred-rgcLSTM and 
PredNet.
The rgcLSTM uses fewer trainable parameters.
Also, 
the
training time is reduced by about $40\%-50\%$ as in the moving MNIST (gray-scale) prediction task and a smaller SE, MSE, and MAE values. 

For the KITTI dataset,
sample visual testing results comparing our rgcLSTM and the cLSTM appear in Figure~\ref{KITTI_results}. 
This figure shows that the rgcLSTM 
(with smaller training time and parameter count)
matches the accuracy
in comparison to the cLSTM. 
To compare the accuracy more clearly between the rgcLSTM and cLSTM, images of prediction step 6 
are magnified
for each model.
This is shown in Figure~\ref{KITTI_zoom}.
The predictions for both models in comparison to ground truth are 
perceptually indistinguisable (although the images have limited resolution).

Figure~\ref{kitti_train_loss} shows the training and validation loss through $150$ epochs of the
KITTI dataset training using both the proposed Pred-rgcLSTM and PredNet. 
The graphs show that the training and validation losses of both models is nearly identical
suggesting that both models may be approaching the limit in Bayesian prediction error. 
The details of the random walk match closely in both models.
This phenomenon is robust as we have observed it in other simulation runs.
It suggests the rgcLSTM captures the nature of the cLSTM computation in great detail.

\subsubsection{Discussion of Experiment 2}
The general pattern of results found in Experiment~1 using the Moving MNIST dataset
was reproduced in our second experiment that used the KITTI dataset.
First, performance was nearly equivalent for both models.
Second, the cost-effectiveness of rgcLSTM-PredNet in relation to PredNet persists.
In particular, the parameter count, memory needed to store the trained weights,
and the training time are all greatly reduced.

\subsection{Experiment 3: Studying the Space of Reduced Models}
The good performance of the rgcLSTM version of PredNet raises the following question.
The three gates of the standard LSTM must serve some purpose.
It seems too good to be true that the one gate of the rgcLSTM could effectively learn to perform the
three different gate functions of a standard LSTM\@,
and thereby act as a multi-function gate.
It suggests that the separate gates of the standard LSTM have excess learning capacity for the Moving
MNIST and KITTI prediction tasks
and that the single set of shared weights of a single gate can learn to perform all three different
functions.
To explore this possibility 
further,
we enumerated a subspace of gated recurrent architectures and examined their performance
on both the Moving MNIST and KITTI datasets.

\subsubsection{Method}
\label{sec_methodForExp3}
We studied two sets of gated networks that included a total of twenty models.
The first was a set of fourteen \textit{single-function} gate models.
The second set consisted of six \textit{multi-function} gate models.

First, we explain how the single-function gate models were chosen.
We started with the full three-gate convLSTM of Shi et al.~\cite{Shi2015} which included elementwise
peephole connections.
In addition to the peephole connections, this architecture has three gates.
We include it in the set.
Next, we obtain three two-gate architectures by removing one of the three gates
from the original convLSTM.\@
Finally, there are three ways to include one gate.
This yields a total of 1+3+3=7 architectures.
For each of these architectures, we can choose to use an elementwise peephole connection, 
or not.
The elementwise version is used because we are considering reductions of the Shi et al.~\cite{Shi2015} model.
This gives $2\cdot 7=14$ architectures.

Next, we explain how the multi-function gate models were chosen.
The goal here is to create reduced models of our rgcLSTM\@.
Like our rgcLSTM, each of these models has exactly one gate.
Because it is a multi-function gate,
this gate is constrained to perform more than one function simultaneously.
In all cases, the gate serves as a forget gate.
The gate can also be used to perform either one or two other functions, yielding three cases:
\begin{enumerate}
\item The gate serves all three functions.
\item The gate serves two functions, specifically a forget gate and also as an input gate.
\item The gate serves two functions, specifically a forget gate and also as an output gate.
\end{enumerate}

\noindent
Finally, we can make the choice to add, or not add, a (convolutional) peephole connection.
The convolutional peephole connection was used because we are considering reductions of
the rgcLSTM model (which used convolutional peephole connections). 
This gives $2\cdot3=6$ architectures.

The twenty resulting architectures are shown in the appendix.
The equations for models M1 through M14 are shown in Table~\ref{non_shared_models}
and describe the single-function gate architectures.
The equations for models M15 through M20 are shown in Table~\ref{models_shared}
and describe the multi-function gate architectures.

M1 is an alias for the cLSTM architecture used in the original Lotter et al.~\citep{Lotter2017} model and M8 is an alias
for the Shi et al.~\citep{Shi2015} convLSTM with elementwise peephole connections.
M18 is an alias for our rgcLSTM architecture having a single gate with three functions as well as
a convolutional peephole connection.

\begin{table*}
	\caption{The single-function gate performance for the oving MNIST datas set.}
	\scriptsize
	\begin{center}
		\begin{tabular}{lllllllllll}
			\multicolumn{1}{l}{\bf Model} &\multicolumn{1}{l}{\bf Peephole}  &\multicolumn{1}{l}{\bf f-gate }&\multicolumn{1}{l}{\bf i-gate}&\multicolumn{1}{l}{\bf o-gate}&\multicolumn{1}{l}{\bf MSE}&\multicolumn{1}{l}{\bf MAE}&\multicolumn{1}{l}{\bf SSIM}&\multicolumn{1}{l}{\bf Memory(MB)}&\multicolumn{1}{l}{\bf \# parameters}
			\\ \hline 
			M1&No& \checkmark&\checkmark&\checkmark& \textbf{0.010}& \textbf{0.019}&\textbf{0.915}&81.068&6,909,834\\
			M2&No&\checkmark&-&-&0.043&0.050&0.623&44.479&3,880,786\\
			M3&No&\checkmark&\checkmark&-&0.043&0.050&0.623&61.822&5,395,310\\
			M4&No&\checkmark&-&\checkmark& 0.012& 0.022&0.890&61.825&5,395,310\\
			M5&No&-&\checkmark&-&0.043&0.050&0.623&44.479&3,880,786\\
			
			M6&No&-&\checkmark&\checkmark&0.043&0.050&0.623&61.822&5,395,310\\
			M7&No&-&-&\checkmark&0.043&0.050&0.623&44.479&3,880,786\\
			
			M8&Yes&\checkmark&\checkmark&\checkmark&\textbf{0.010}&\textbf{0.019}&\textbf{0.914}&94.121&8,216,229\\
			M9&Yes&\checkmark&-&-&0.043&0.050&0.623&49.461&4,316,251\\
			M10&Yes&\checkmark&\checkmark&-&0.013&0.023&0.881&71.790&6,266,240\\
			M11&Yes&\checkmark&-&\checkmark&0.043&0.050&0.623&71.791&6,266,240\\
			M12&Yes&-&\checkmark&-&0.043&0.050&0.623&49.463&4,316,251\\
			M13&Yes&-&\checkmark&\checkmark&0.043&0.050&0.623&71.791&6,266,240\\
			M14&Yes&-&-&\checkmark&0.020&0.032&0.813&49.463&4,316,251\\
			
			\hline
		\end{tabular}
		\label{nonshared_mnist}
	\end{center}
\end{table*}

\begin{table*}
	\caption{The multi-function gate performance for the Moving MNIST dataset.}
	\scriptsize
	\begin{center}
		\begin{tabular}{ llllllllll}
			\multicolumn{1}{l}{\bf Model} &\multicolumn{1}{l}{\bf Peephole}  &\multicolumn{1}{l}{\bf Acting Gate}&\multicolumn{1}{l}{\bf Shared Effect}&\multicolumn{1}{l}{\bf MSE}&\multicolumn{1}{l}{\bf MAE}&\multicolumn{1}{l}{\bf SSIM}&\multicolumn{1}{l}{\bf Memory(MB)}&\multicolumn{1}{l}{\bf \# parameters}
			\\ \hline 
			M15&No& f-gate& i-gate, o-gate&\textbf{0.009}& \textbf{0.018}& \textbf{0.918}&44.479&3,880,786 \\
			M16&No&f-gate& i-gate&0.043&0.050&0.623&44.479&3,880,786\\
			M17&No&f-gate& o-gate&0.011&0.020&0.908&44.479&3,880,786\\			
			M18&Yes& f-gate& i-gate, o-gate&\textbf{0.009}&\textbf{0.017}&\textbf{0.925}&50.650&4,316,235\\
			M19&Yes& f-gate& i-gate&0.013&0.024&0.879&50.650&4,316,235\\
			M20&Yes& f-gate& o-gate& 0.015&0.025&0.860&50.650&4,316,235\\
			
			\hline
		\end{tabular}
		\label{shared}
	\end{center}
\end{table*}

\begin{figure*}
	\centering
	\includegraphics[width=9.34cm, height=6cm]{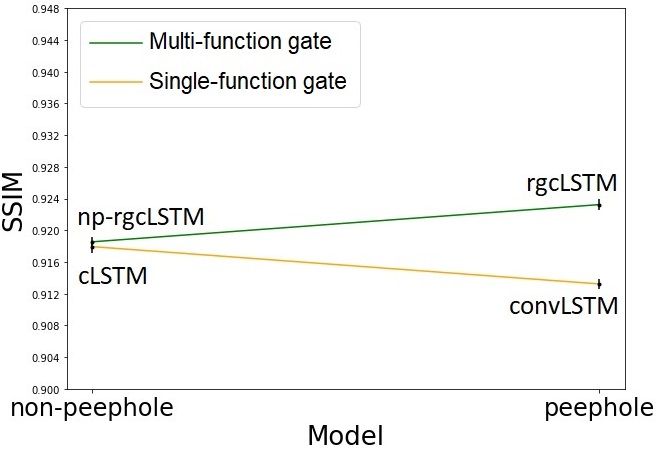}
	\caption{The SSIM for the single-function and multi-function gate architectures for the Moving MNIST dataset video prediction.}
	\label{err_plot_mnist}
\end{figure*}

\subsubsection{Results for the Moving MNIST Data Set}
This section reports the results for all twenty models on the Moving MNIST dataset.
The following section will report results for the KITTI dataset.
Table~\ref{nonshared_mnist} 
describes the single-function gate models and shows the mean MSE, MAE, and SSIM\@.
The means are based on fifteen simulation runs.
Also shown is the memory consumed by each model and the parameter count.
By all three performance measures (MSE, MAE, and SSIM), models M1 (cLSTM) and M8 (convLSTM) 
gave the best performance.
Of these two models, M1 had the lower memory requirement and parameter count.

Table~\ref{shared} reports the performance for the multi-function gate models.
Models M15 and M18, which each use a 3-function gate, perform better than the other models
which use a 2-function gate.
Recall that M18 is an alias for our rgcLSTM and M15 is a non-peephole version of M18, the np-rgcLSTM.

The single-gate models M2, M5, M7, M9, M12, and M14 from Table~\ref{nonshared_mnist} can be meaningfully compared
with models M16, M17, and M18 in Table~\ref{shared}.
These models are also multi-function gate models in the degenerate case
where the gate performs only one function.
With the exception of M14, their performance is the same as the worst performing 2--function gate model
in Table~\ref{shared}.
This reflects the trend that when the number of gate functions is reduced from three to two to one, performance
is reduced.

\begin{table*}
	\caption{Total number of observations, mean SSIM, standard deviation, standard error, and the 95\% confidence interval grouped by the models with respect to the SSIM over the moving MNIST prediction.}
	\begin{center}
		\begin{tabular}{lcrlllr }
			\multicolumn{1}{l}{\bf Model} &\multicolumn{1}{l}{\bf N}  &\multicolumn{1}{l}{\bf Mean SSIM}&\multicolumn{1}{l}{\bf SD}&\multicolumn{1}{l}{\bf SE}&\multicolumn{1}{l}{\bf 95\% Conf. Interval}
			\\ \hline 
			cLSTM     &  15 &0.915 & 0.00156 & 0.00040  & 0.9171-0.9187\\
			convLSTM  &  15 &0.913 & 0.00124 & 0.00032  & 0.9126-0.9139\\
			np-rgcLSTM&  15 &0.919 & 0.00134 & 0.00035  & 0.9179-0.9190\\
			rgcLSTM   &  15 &0.924 & 0.00144 & 0.00037  & 0.9225-0.9240\\
			\hline
		\end{tabular}
		\label{anova_ssim_mnist_table2}
	\end{center}
\end{table*}

\begin{table*}
	\caption{Post-hoc testing over the four models shown in Figure~\ref{err_plot_mnist} for the Moving MNIST video prediction task.}
	\begin{center}
		\begin{tabular}{llrrrr}
			\multicolumn{1}{l}{\bf Group1}&\multicolumn{1}{l}{\bf Group2} &\multicolumn{1}{l}{\bf MeanDiff}  &\multicolumn{1}{l}{\bf Lower}&\multicolumn{1}{l}{\bf Upper}&\multicolumn{1}{l}{\bf Reject}
			\\ \hline 
			cLSTM &   np-rgcLSTM &   0.0006 & -0.0007 & 0.002 & False \\
			cLSTM &   rgcLSTM &   0.0053 &  0.0039 & 0.0067&  True \\
			cLSTM   &  convLSTM &  -0.0047 & -0.0061& -0.0034&  True \\
			np-rgcLSTM   &  rgcLSTM &   0.0047 &  0.0033 & 0.006 &  True \\
			np-rgcLSTM   &   convLSTM &  -0.0053 & -0.0067 & -0.004&  True \\
			rgcLSTM  &   convLSTM  &  -0.01  & -0.0114 &-0.0087&  True \\
			\hline
		\end{tabular}
		\label{post_hoc_mnist}
	\end{center}
\end{table*}

The SSIM performance of the four winning models (M1, M8, M15, and M18) is plotted in Figure~\ref{err_plot_mnist}.
The plot is broken down by presence or absence of a peephole connection and whether the model used single-function
or multi-function gates.
95\% confidence interval error bars are displayed on the figure and were calculated according to 
Table~\ref{anova_ssim_mnist_table2}.
A two-way analysis of variance revealed that there was a significant interaction (p $< .05$) between the 
presence/absence of peephole connections and model type (single-function versus multi-function gate).
The figure shows that the peephole connection reduced performance of the cLSTM but improved performance of the rgcLSTM\@.
The post-hoc comparison in Table~\ref{post_hoc_mnist} shows that all comparisons are different except for the cLSTM versus
np-rgcLSTM, which is easily seen in Figure~\ref{err_plot_mnist}.
We will postpone further discussion until after we report the results for the KITTI dataset.

\subsubsection{Results for the KITTI Data Set}
\begin{figure*}
	\centering
	\includegraphics[width=9.34cm, height=6cm]{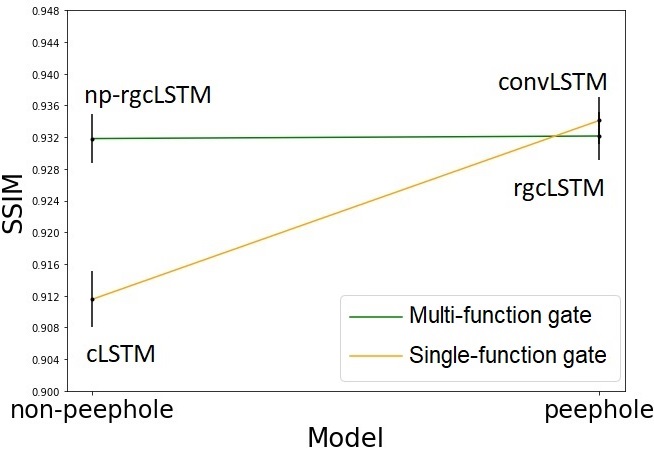}
	\caption{The SSIM for the single-function and multi-function gate architectures for the KITTI dataset video prediction.}
	\label{err_plot_kitti}
\end{figure*}

We selected the four winning Moving MNIST models (M1, M8, M15, M18) for comparative study on the
KITTI dataset.
Figure~\ref{err_plot_kitti} shows the SSIM performance results for the four models with 95\% confidence intervals.
Qualitatively, the pattern of results differs from those shown in Figure~\ref{err_plot_mnist} for the
Moving MNIST dataset.
This time the peephole connections improve the performance of the cLSTM but have no effect for the rgcLSTM\@.
A two-way ANOVA reveals that the interaction is significant (p $< .05$).
The post-hoc comparisons in Table~\ref{post_hoc_kitti} indicate that the cLSTM performs worse than the other three models
on the KITTI dataset.

\begin{table*}
	\caption{Post-hoc testing over the four models shown in Figure~\ref{err_plot_kitti} for KITTI video prediction task.}
	\begin{center}
		\begin{tabular}{llllllll }
			\multicolumn{1}{l}{\bf Group1}&\multicolumn{1}{l}{\bf Group2} &\multicolumn{1}{l}{\bf MeanDiff}  &\multicolumn{1}{l}{\bf Lower}&\multicolumn{1}{l}{\bf Upper}&\multicolumn{1}{l}{\bf Reject}
			\\ \hline 
			cLSTM   & np-rgcLSTM   & 0.0203 &  0.0119& 0.0286 & True \\
			cLSTM     &rgcLSTM &   0.0206  & 0.0123& 0.0289 & True \\
			cLSTM   &   convLSTM   & 0.0226  & 0.0142 &0.0309 & True \\
			np-rgcLSTM  &  rgcLSTM  &  0.0003  & -0.008& 0.0087& False \\
			np-rgcLSTM   &  convLSTM &   0.0023  & -0.006 &0.0106& False \\
			rgcLSTM   &  convLSTM &   0.002  & -0.0064& 0.0103& False \\
			\hline
		\end{tabular}
		\label{post_hoc_kitti}
	\end{center}
\end{table*}

\subsubsection{Discussion of Experiment 3}
This experiment was performed to explore the space of single-function and multi-function
gated models in an effort to resolve the question of how our rgcLSTM can successfully use one gate to perform
functions that normally use three gates, as in the cLSTM\@.
We found that when these models were substituted into Lotter et al.'s PredNet on the Moving MNIST dataset, 
the best performing 
models included our rgcLSTM and Lotter et al.'s cLSTM, according to the MSE, MAE, and SSIM performance
measures.
When broken down by two factors: 1) whether or not peephole connections were used; and 2) whether
single-function or multi-function gates were used,
the best performing models for each of these conditions were cLSTM, convLSTM, np-rgcLSTM, and rgcLSTM
as shown in Tables~\ref{nonshared_mnist}, \ref{shared}, and~\ref{anova_ssim_mnist_table2}.
This result is also visualized in Figure~\ref{err_plot_mnist} and corroborates Experiment~1 where the 
rgcLSTM performed better than the cLSTM\@.
These results confirm that the rgcLSTM performance makes empirical sense, which was the original
motivation for Experiment 3.
This is seen in Table~\ref{shared} and in the performance the single-gate models M2, M5, M7, M9,
and M12.
For a single gate, as the number of gate functions decreases from three to one, the performance
decreases.

When we studied these four models on the KITTI dataset, we still found that the rgcLSTM
performed best but also that the performance of the convLSTM of Shi et al.~\citep{Shi2015} performed better 
than the cLSTM used by Lotter et al.
This result is specific to the KITTI dataset and was not found for the Moving MNIST dataset.
The other conclusion from this experiment is that the pattern of relative performance of these
models varies with dataset.

\section{Conclusion}
The novelty of our
new
rgcLSTM architecture lies in the following 
aspects.
First, there is one 
multi-function
gate which serves the function of the forget, input, and output gates.
Because only one gate is used,
this reduces the number of gate parameters to one third that of 
the
cLSTM
or convLSTM.
Second, in this reduced
parameter, multi-function gate
model there is still a peephole connection from the cell memory state
to the module gate which allows the cell memory 
to
maintain 
control over the module gate, in contrast to the cLSTM\@.
Finally, the rgcLSTM uses a convolutional architecture and,
because of this,
we have replaced the elementwise
multiply 
peephole
operation originally used in Shi et al.~\citep{Shi2015} with a convolutional
peephole connection.
Because the peephole connection is convolutional, the associated weight sharing, 
reduces the parameter count that would be required if an elementwise peephole connection
was used.
This change deviates from the convLSTM, yet is still compatible with the image processing
needs of a convolutional LSTM\@.
Our experiments show that
these 
changes support
the model's ability for 
state-of-the-art video prediction.
Thus, 
the 
proposed rgcLSTM model reduces the number of trainable parameters, training time, and memory requirements.

The results of Experiments~1 and~2 were consistent with each other and did not reveal any difference
in how the rgcLSTM-PredNet model processed the datasets.
A more detailed study was performed in Experiment~3 involving a large set of possible
reduced parameter, single-function gate architectures and multi-function gate architectures.
It was shown that the PredNet variants using the rgcLSTM, cLSTM, and convLSTM modules performed
best on the Moving MNIST dataset.
This confirms the effectiveness or our rgcLSTM as well as the earlier choice of the cLSTM in the literature.
It also confirms that the convLSTM, not used by Lotter et al., would also have been a good choice.
Experiment~3 also reveals performance differences between the Moving MNIST and KITTI datasets in
regard to choice of architecture.
Specifically, the rgcLSTM performed best on the Moving MNIST dataset
and peephole connections  reduced performance when using single-function gate convolutional LSTMs.
On the KITTI dataset, the pattern of results was different.
Including peephole connections had no effect when using an rgcLSTM, but yielded a significant
improvement when using a convolutional LSTM\@.

For future work, 
experiments are needed on a larger sample of datasets to see whether there is some kind of consistent
performance pattern across datasets.
Experiments are also needed that use more challenging datasets to give the rgcLSTM model a stress test
to see when and if its multi-function gates are overwhelmed by a more challenging dataset.
Finally, the gated recurrent unit (GRU) model was the first multi-function gate model used in
the literature. 
Future experiments should compare the GRU with the rgcLSTM\@.
Although Experiment~3 studied six different multi-function gate models, it omitted the GRU
because it did not fit within our enumeration scheme.

Despite this parameter reduction,
the proposed rgcLSTM model 
performs at a state-of-the-art level on the datasets studied.
Because of the reduced parameter count, memory footprint, and training time,
our rgcLSTM model 
is an
attractive 
candidate
for future hardware implementation on small and mobile devices.

\clearpage
\section{Appendix}
Tables~\ref{non_shared_models} and~\ref{models_shared} show the equations and cell diagrams
for the models used in Experiment~3.
Table~\ref{non_shared_models} describes the single-function gate models.
Table~\ref{models_shared} describes the multi-function gate models.

	\begin{longtable}{p{1.0cm} p{6.5cm} p{9.6cm}}
	\renewcommand{\arraystretch}{0.7}\\
	\caption{The fourteen single-function gate convolutional LSTM variants. M1 is an alias for the cLSTM~\cite{Lotter2017} and M8 is an alias for the convLSTM~\cite{Shi2015}.}\\
	\textbf{\textit{Model}} &\textbf{\textit{Diagram}}& \textbf{\textit{Formulas}}\\
	\\ \hline
	\\
	M1&  \begin{minipage}{6cm}
		\includegraphics[width=4.8cm, height=4cm]{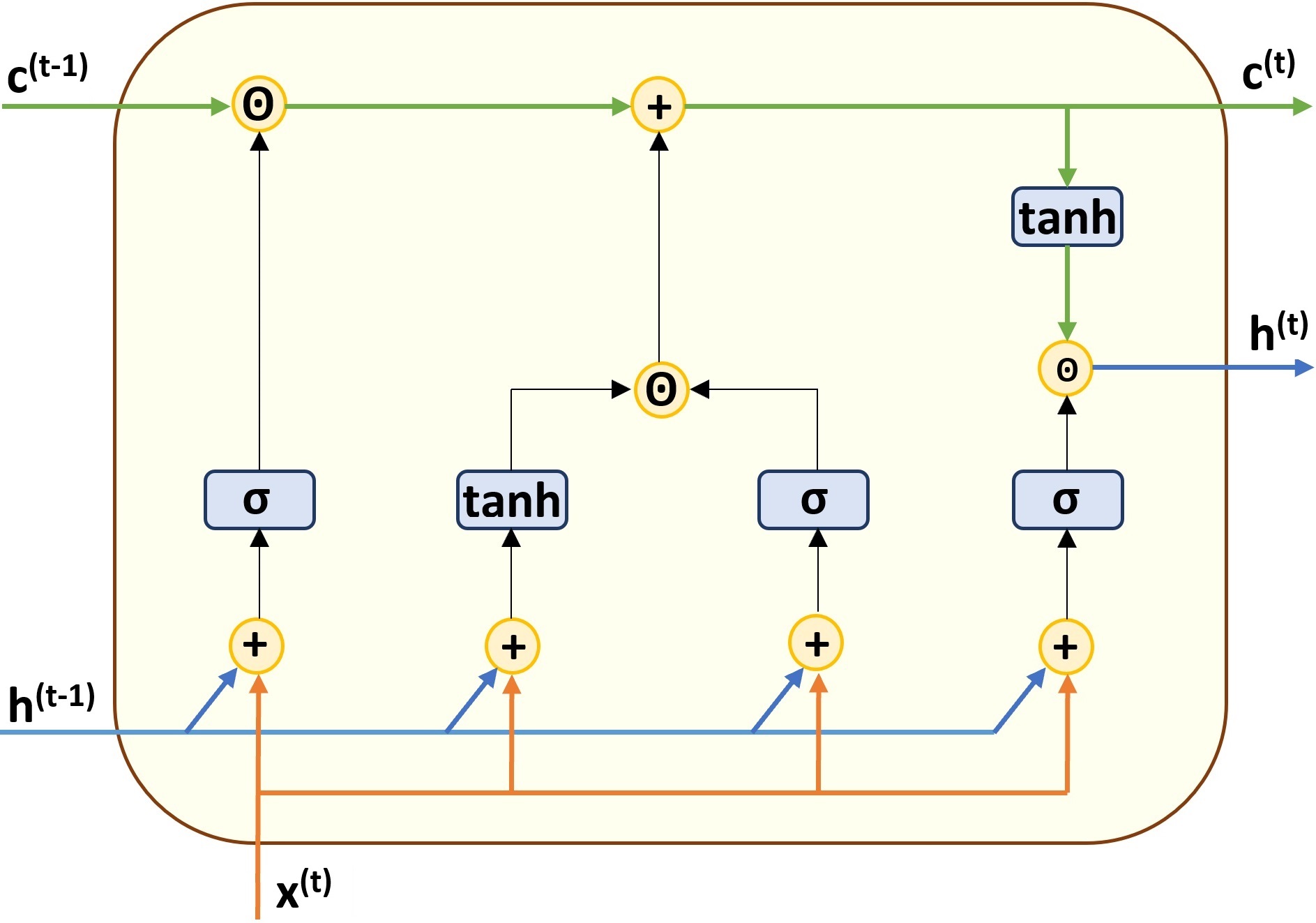}
	\end{minipage} & 
	\parbox{8.5cm}{\begin{align}
		f^{(t)}&=\sigma([W_{fx},U_{fh}]*[x^{(t)}, h^{(t-1)}] + b_f)\label{eqn1}\\ 
		i^{(t)}&= \sigma([W_{ix},U_{ih}] * [x^{(t)}, h^{(t-1)}]+ b_i)\label{eqn2}\\ 
		o^{(t)}&=\sigma([W_{ox},U_{oh}] * [x^{(t)}, h^{(t-1)}] + b_o)\label{eqn3}\\ 
		g^{(t)}&= tanh([W_{gx},U_{gh}] *[x^{(t)}, h^{(t-1)}]+ b_g)\label{eqn4}\\ 
		c^{(t)}&= f^{(t)}\odot c^{(t-1)} + i^{(t)} \odot g^{(t)}\label{eqn5}\\
		h^{(t)}&= tanh(c^{(t)})\odot o^{(t)}\label{eqn6}
		\end{align}}\\
	\\
	M2&  \begin{minipage}{6cm}
		\includegraphics[width=4.8cm, height=4cm]{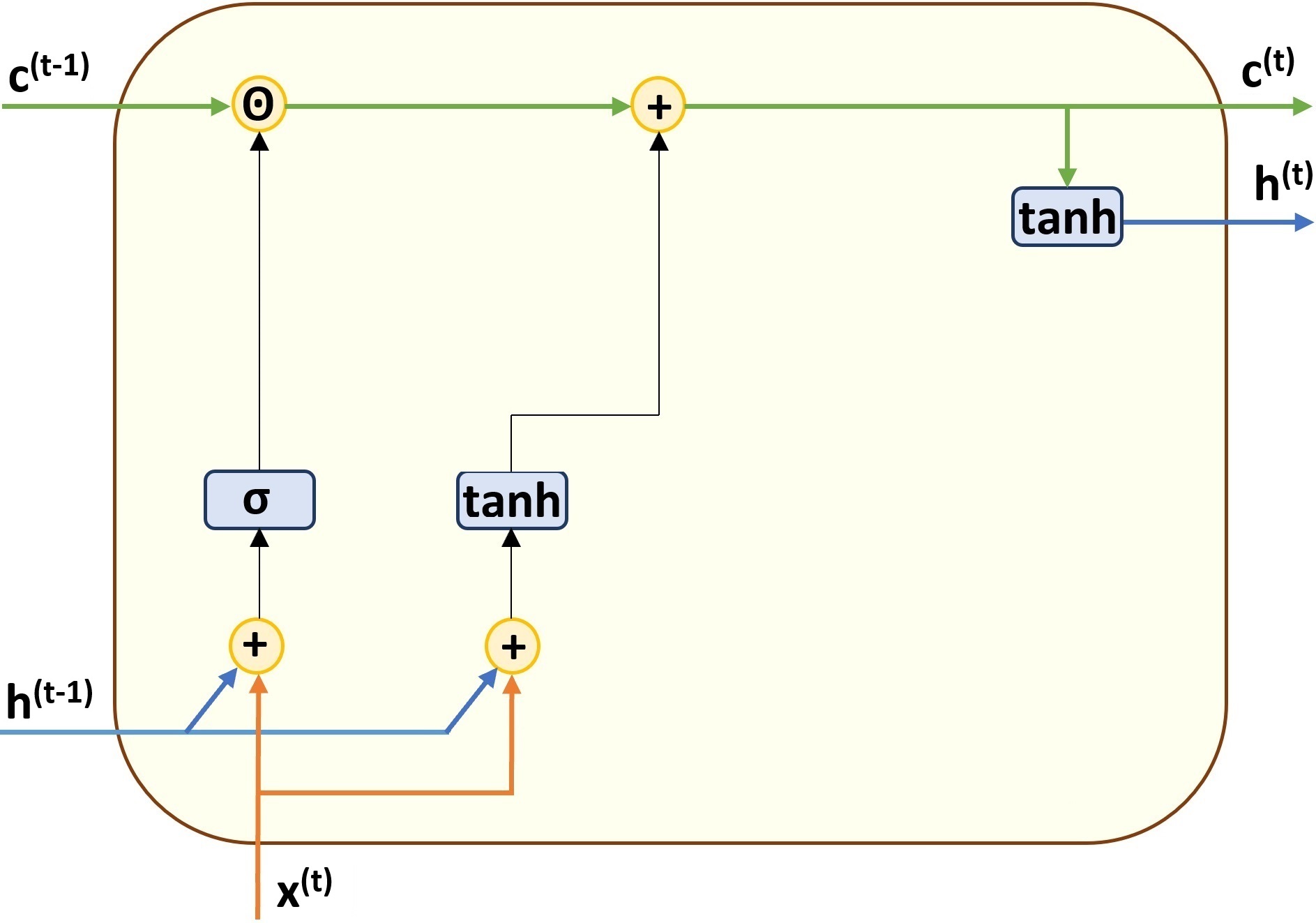}
	\end{minipage} & 
	\parbox{8.5cm}{\begin{align}
		f^{(t)}&=\sigma([W_{fx},U_{fh}] * [x^{(t)}, h^{(t-1)}] + b_f)\label{eqn7}\\ 
		g^{(t)}&= tanh([W_{gx},U_{gh}] * [x^{(t)}, h^{(t-1)}]+ b_g)\label{eqn8}\\ 
		c^{(t)}&= f^{(t)}\odot c^{(t-1)} + g^{(t)}\label{eqn9}\\
		h^{(t)}&= tanh(c^{(t)})\label{eqn10}
		\end{align}}\\
	\\
	
	M3&  \begin{minipage}{6cm}
		\includegraphics[width=4.8cm, height=4cm]{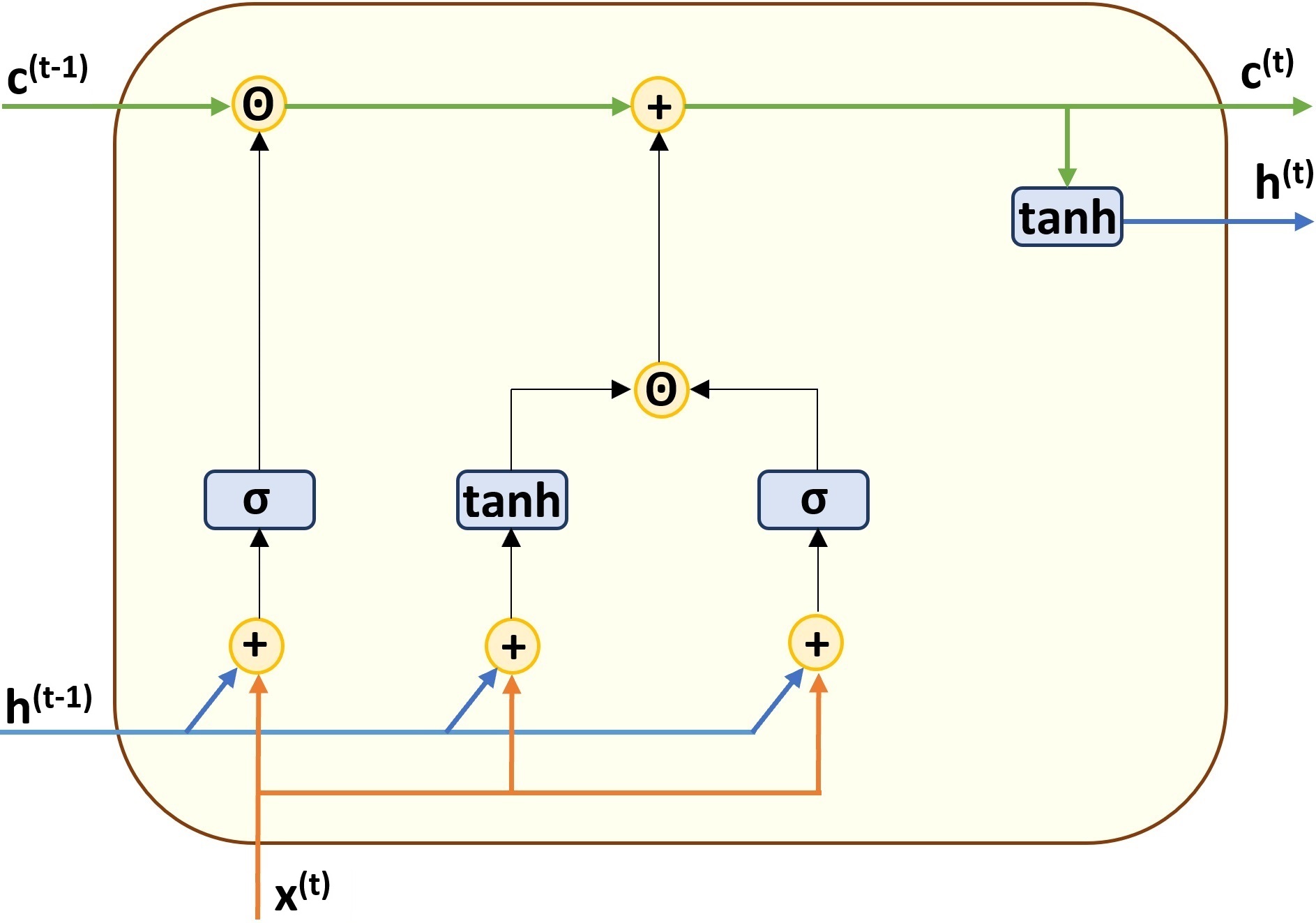}
	\end{minipage} & 
	\parbox{8.5cm}{\begin{align}
		f^{(t)}&=\sigma([W_{fx},U_{fh}] *[x^{(t)}, h^{(t-1)}] + b_f)\label{eqn11}\\ 
		i^{(t)}&= \sigma([W_{ix}, +U_{ih}] * [x^{(t)}, h^{(t-1)}]+ b_i)\label{eqn12}\\ 
		g^{(t)}&= tanh([W_{gx},U_{gh}] *[x^{(t)}, h^{(t-1)}]+ b_g)\label{eqn13}\\ 
		c^{(t)}&= f^{(t)}\odot c^{(t-1)} + i^{(t)} \odot g^{(t)}\label{eqn14}\\
		h^{(t)}&= tanh(c^{(t)})\label{eqn15}
		\end{align}}\\
	\\
	M4&  \begin{minipage}{6cm}
		\includegraphics[width=4.8cm, height=4cm]{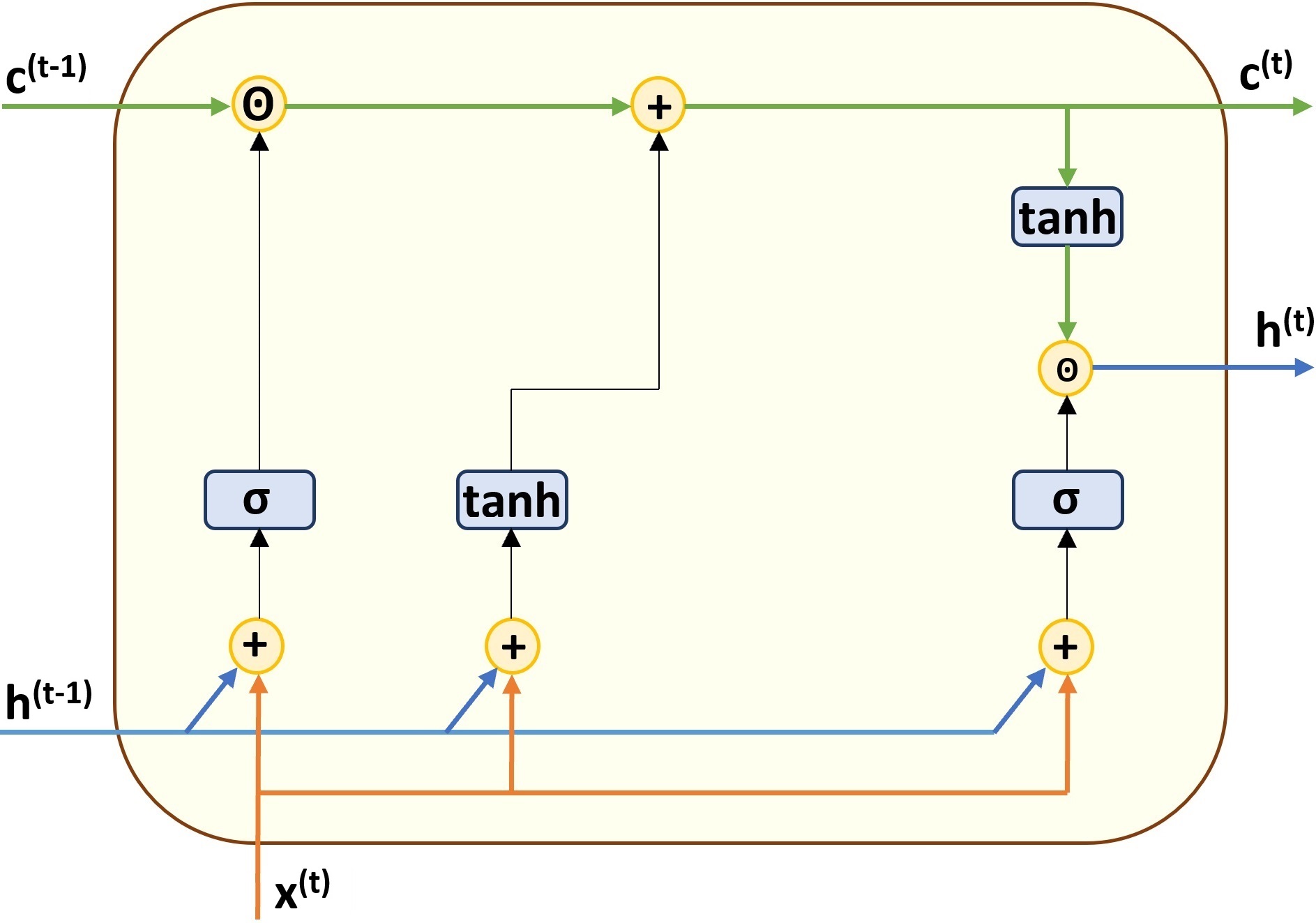}
	\end{minipage} & 
	\parbox{8.5cm}{\begin{align}
		f^{(t)}&=\sigma([W_{fx},U_{fh}] *[x^{(t)}, h^{(t-1)}] + b_f)\label{eqn16}\\ 
		o^{(t)}&=\sigma([W_{ox},U_{oh}] * [x^{(t)}, h^{(t-1)}]+ b_o)\label{eqn17}\\ 
		g^{(t)}&= tanh([W_{gx},U_{gh}] * [x^{(t)}, h^{(t-1)}]+ b_g)\label{eqn18}\\ 
		c^{(t)}&= f^{(t)}\odot c^{(t-1)} + i^{(t)} \odot g^{(t)}\label{eqn19}\\
		h^{(t)}&= tanh(c^{(t)})\odot o^{(t)}\label{eqn20}
		\end{align}}\\
	\\
	M5&  \begin{minipage}{7cm}
		\includegraphics[width=4.8cm, height=4cm]{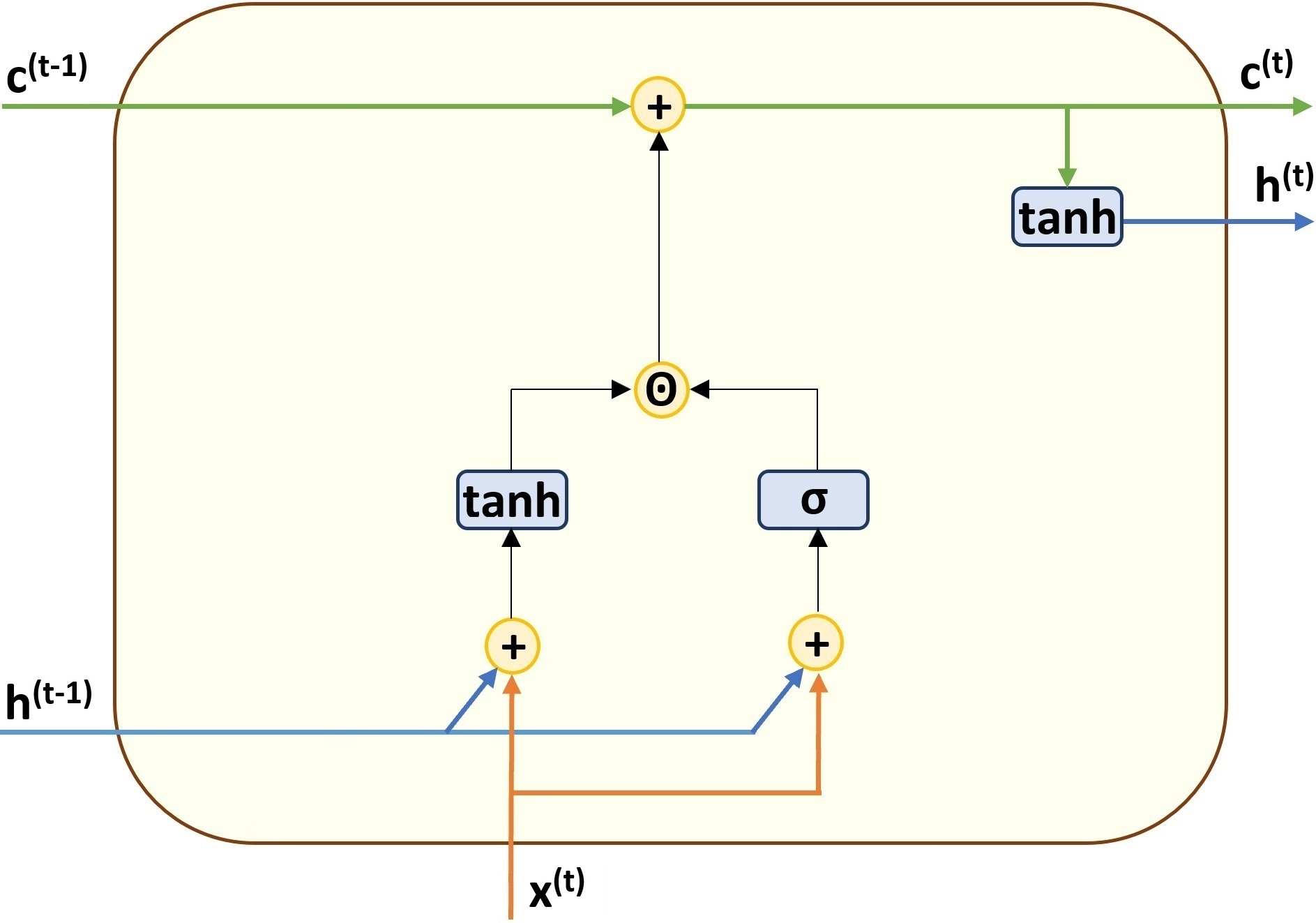}
	\end{minipage} & 
	\parbox{8.5cm}{\begin{align}
		i^{(t)}&= \sigma([W_{ix},U_{ih}] * [x^{(t)}, h^{(t-1)}]+ b_i)\label{eqn21}\\  
		g^{(t)}&= tanh([W_{gx},U_{gh}] * [x^{(t)}, h^{(t-1)}]+ b_g)\label{eqn22}\\ 
		c^{(t)}&= c^{(t-1)} + i^{(t)} \odot g^{(t)}\label{eqn23}\\
		h^{(t)}&= tanh(c^{(t)})\label{eqn24}
		\end{align}}\\
	\\
	M6&  \begin{minipage}{7cm}
		\includegraphics[width=4.8cm, height=4cm]{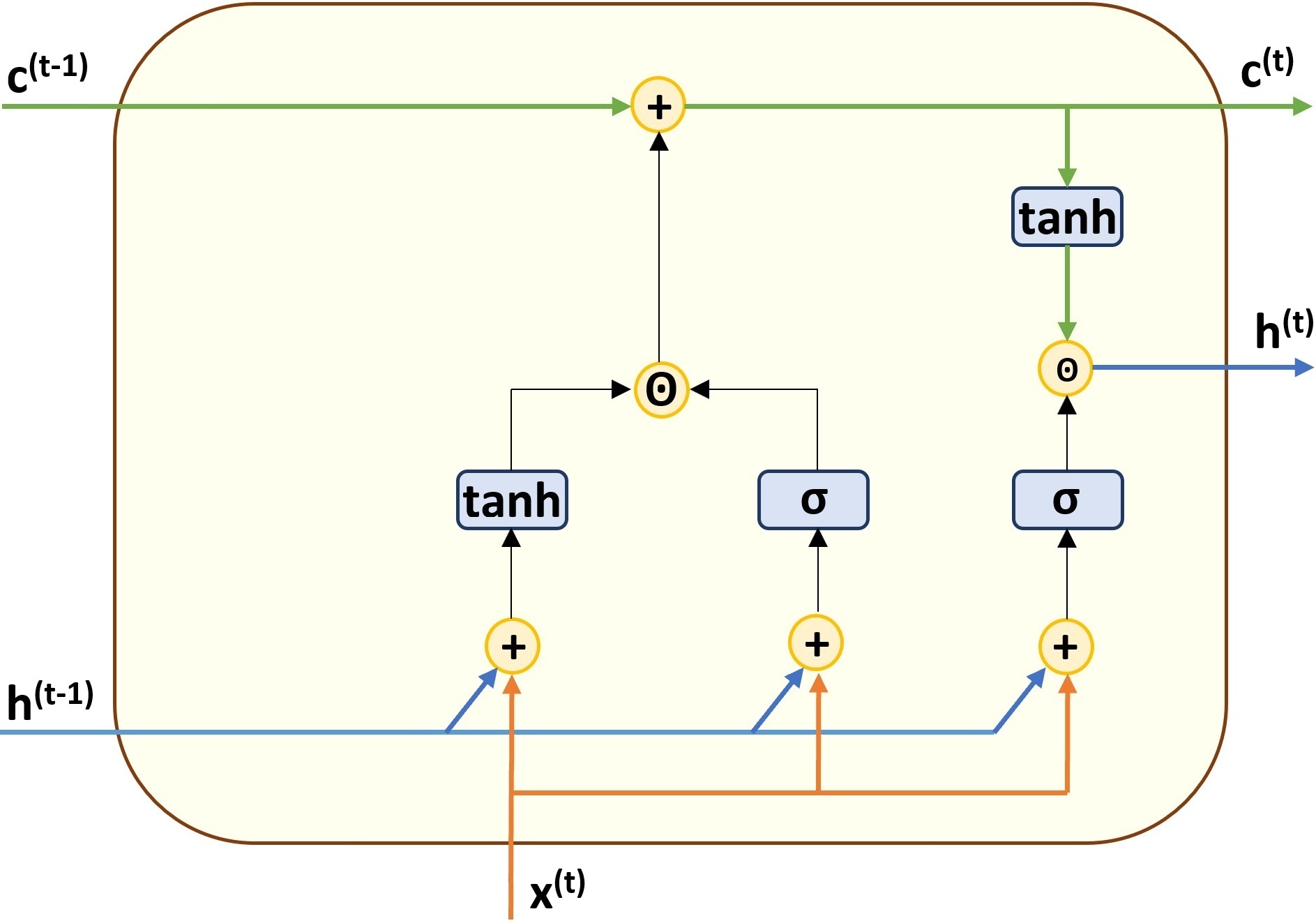}
	\end{minipage} & 
	\parbox{8.5cm}{\begin{align}
		i^{(t)}&= \sigma([W_{ix},U_{ih}] * [x^{(t)}, h^{(t-1)}]+ b_i)\label{eqn25}\\ 
		o^{(t)}&=\sigma([W_{ox},U_{oh}] * [x^{(t)}, h^{(t-1)}] + b_o)\label{eqn26}\\ 
		g^{(t)}&= tanh([W_{gx},U_{gh}] * [x^{(t)}, h^{(t-1)}]+ b_g)\label{eqn27}\\= 
		c^{(t)}&= c^{(t-1)} + i^{(t)} \odot g^{(t)}\label{eqn28}\\
		h^{(t)}&= tanh(c^{(t)})\odot o^{(t)}\label{eqn29}
		\end{align}}\\
	\\
	M7&  \begin{minipage}{6cm}
		\includegraphics[width=4.8cm, height=4cm]{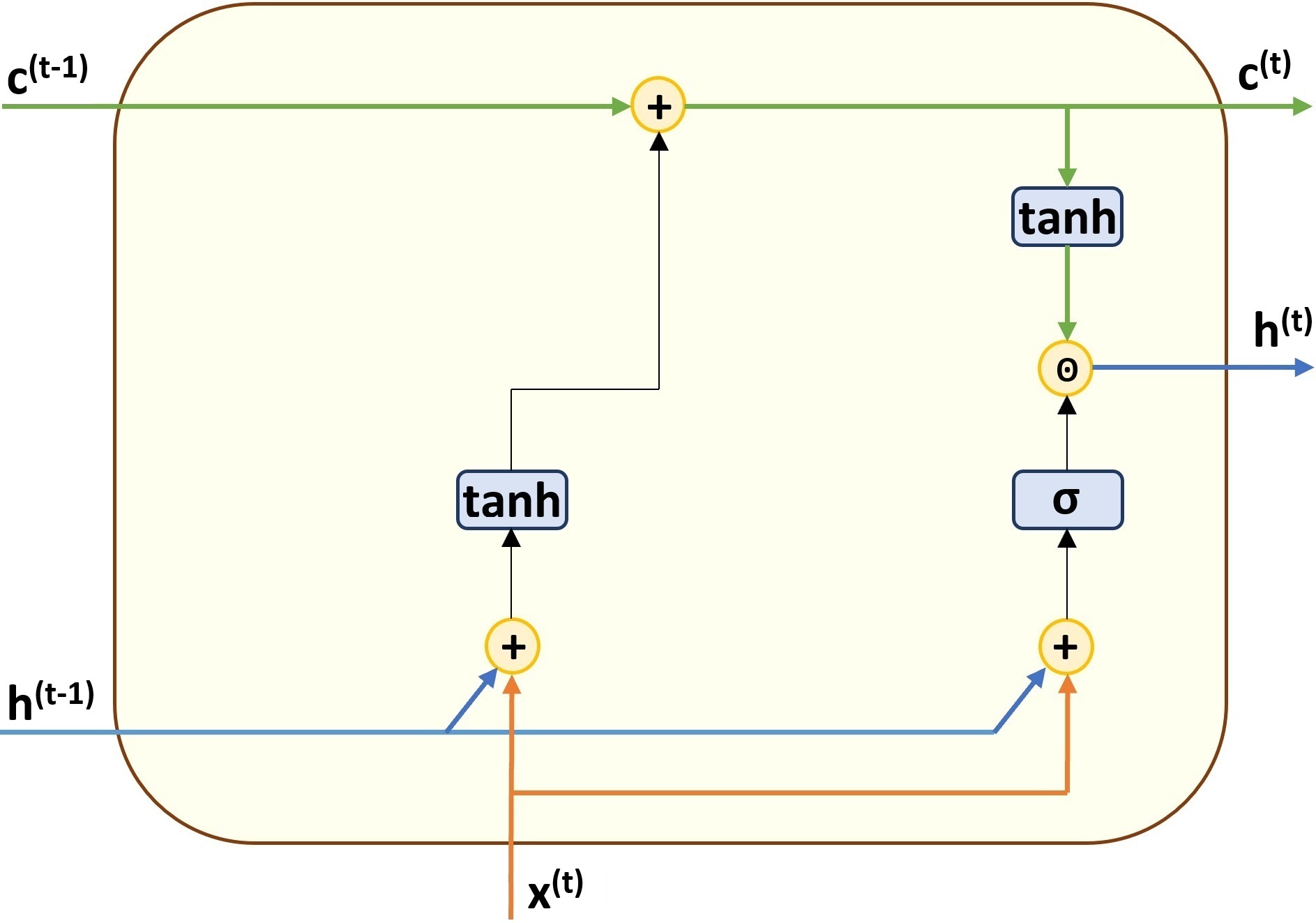}
	\end{minipage} & 
	\parbox{8.5cm}{\begin{align}
		o^{(t)}&=\sigma([W_{ox},U_{oh}] *[x^{(t)}, h^{(t-1)}] + b_o)\label{eqn30}\\ 
		g^{(t)}&= tanh([W_{gx},U_{gh}] * [x^{(t)}, h^{(t-1)}]+ b_g)\label{eqn31}\\ 
		c^{(t)}&= c^{(t-1)} + i^{(t)} \odot g^{(t)}\label{eqn32}\\
		h^{(t)}&= tanh(c^{(t)})\odot o^{(t)}\label{eqn33}
		\end{align}}\\
	\\
	M8&  \begin{minipage}{6cm}
		\includegraphics[width=4.8cm, height=4cm]{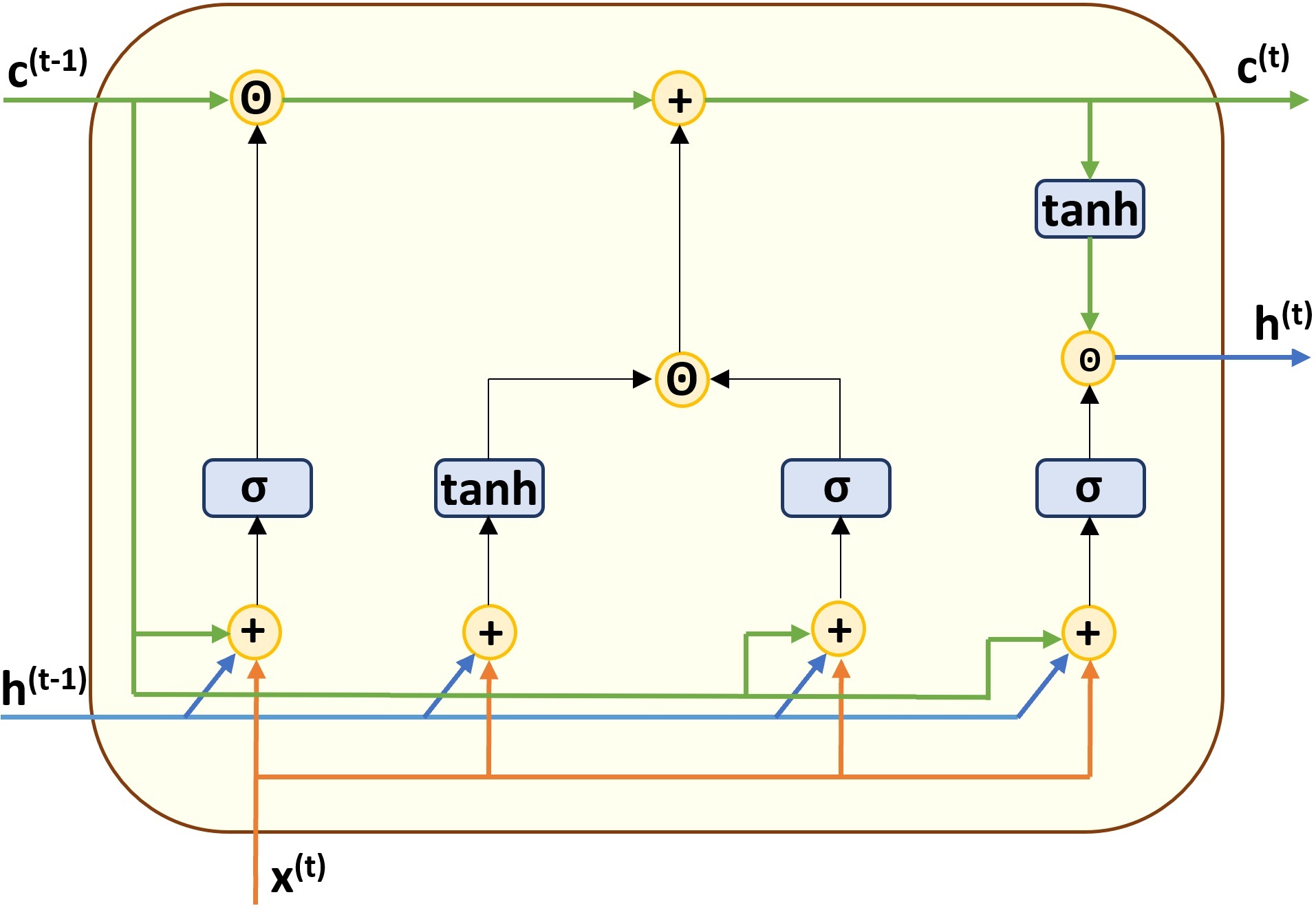}
	\end{minipage} & 
	\parbox{8.5cm}{\begin{align}
		f^{(t)}&=\sigma([W_{fx},U_{fh},W_{fc}] * [x^{(t)}, h^{(t-1)},c^{(t-1)}] + b_f)\label{eqn34}\\
		i^{(t)}&= \sigma([W_{ix},U_{ih},W_{ic}]*[x^{(t)}, h^{(t-1)},c^{(t-1)}]  + b_i)\label{eqn35}\\ 
		o^{(t)}&=\sigma([W_{ox},U_{oh},W_{oc}]*[x^{(t)}, h^{(t-1)},c^{(t-1)}] + b_o)\label{eqn36}\\  
		g^{(t)}&= \mathrm{tanh}([W_{gx},U_{gh}] *[x^{(t)}, h^{(t-1)}] + b_g)\label{eqn37}\\ 
		c^{(t)}&= f^{(t)}\odot c^{(t-1)} + i^{(t)} \odot g^{(t)}\label{eqn38}\\
		h^{(t)}&= \mathrm{tanh}(c^{(t)})\odot o^{(t)}\label{eqn39}
		\end{align}}\\
	\\
	M9&  \begin{minipage}{6cm}
		\includegraphics[width=4.8cm, height=4cm]{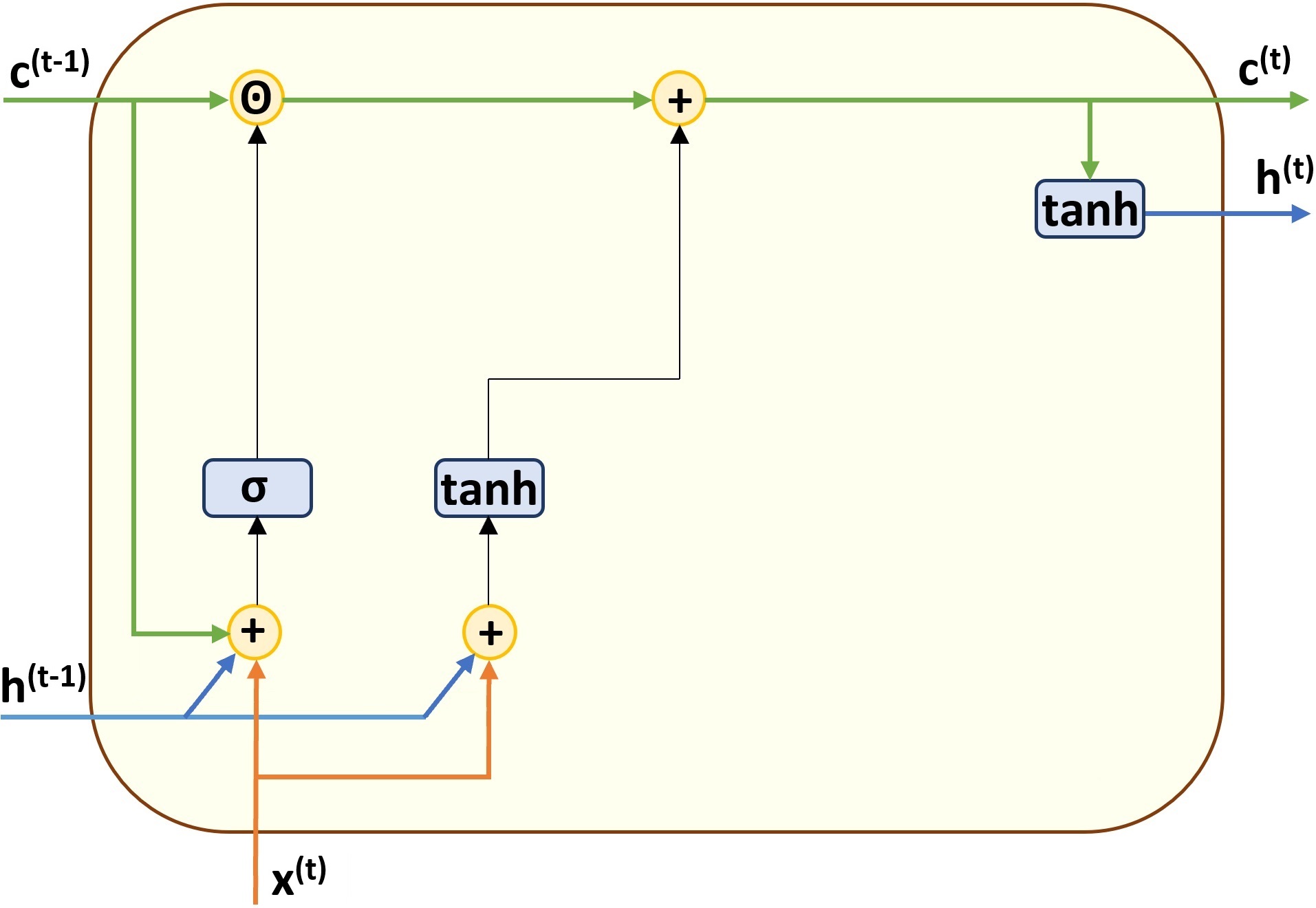}
	\end{minipage} & 
	\parbox{8.5cm}{\begin{align}
		f^{(t)}&=\sigma([W_{fx},U_{fh},W_{fc}] * [x^{(t)}, h^{(t-1)},c^{(t-1)}] + b_f)\label{eqn40}\\ 
		g^{(t)}&= \mathrm{tanh}([W_{gx},U_{gh}] * [x^{(t)}, h^{(t-1)}] + b_g)\label{eqn41}\\ 
		c^{(t)}&= f^{(t)}\odot c^{(t-1)} +  g^{(t)}\label{eqn42}\\
		h^{(t)}&= \mathrm{tanh}(c^{(t)})\label{eqn43}
		\end{align}}\\
	\\
	M10&  \begin{minipage}{6cm}
		\includegraphics[width=4.8cm, height=4cm]{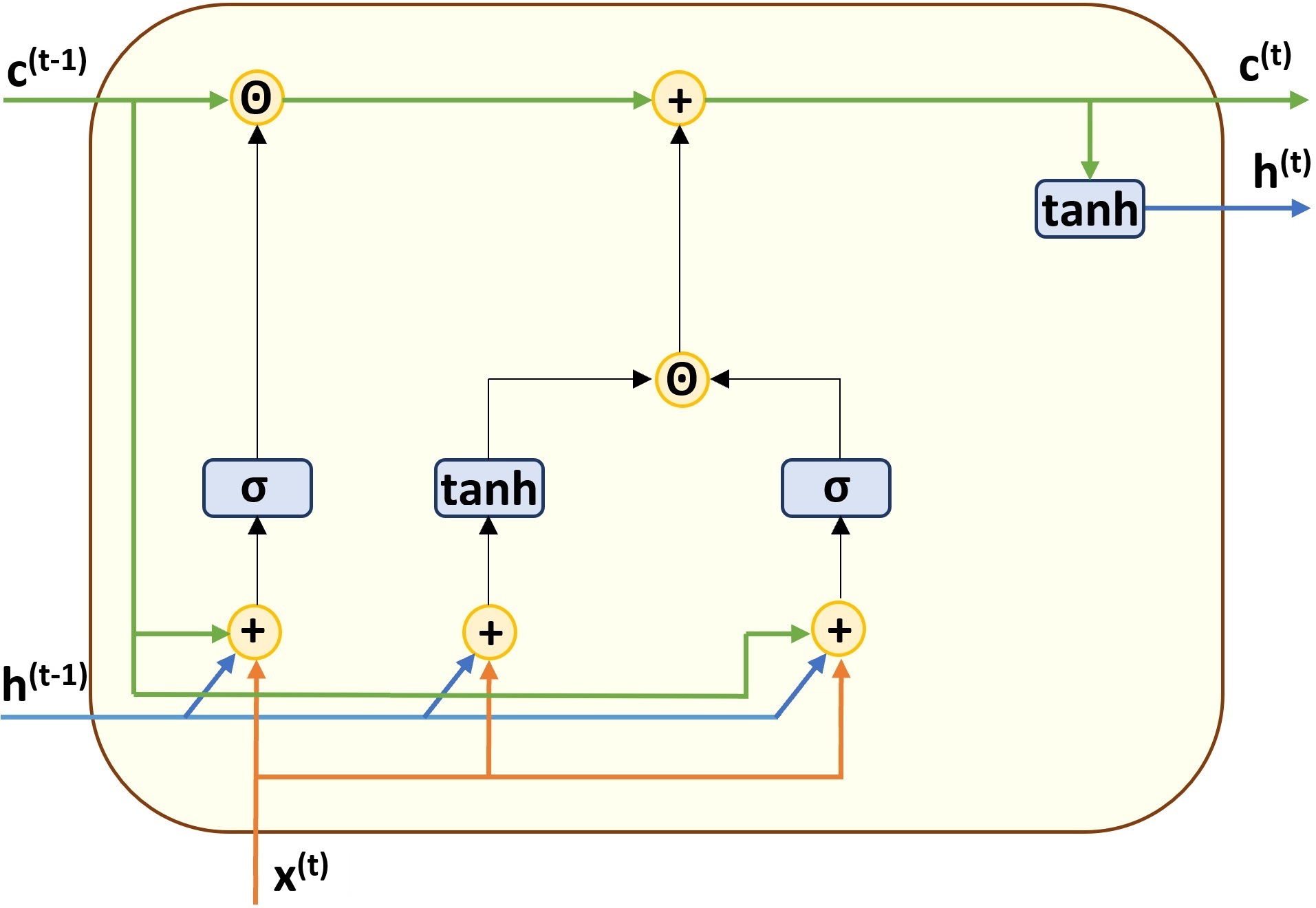}
	\end{minipage} & 
	\parbox{8.5cm}{\begin{align}
		f^{(t)}&=\sigma([W_{fx},U_{fh},W_{fc}] * [x^{(t)}, h^{(t-1)},c^{(t-1)}] + b_f)\label{eqn44}\\ 
		i^{(t)}&= \sigma([W_{ix},U_{ih},W_{ic}]* [x^{(t)}, h^{(t-1)},c^{(t-1)}] + b_i)\label{eqn45}\\ 
		g^{(t)}&= \mathrm{tanh}([W_{gx},U_{gh}] * [x^{(t)}, h^{(t-1)}] + b_g)\label{eqn46}\\ 
		c^{(t)}&= f^{(t)}\odot c^{(t-1)} + i^{(t)} \odot g^{(t)}\label{eqn47}\\
		h^{(t)}&= \mathrm{tanh}(c^{(t)})\label{eqn48}
		\end{align}}\\
	\\
	M11&  \begin{minipage}{6cm}
		\includegraphics[width=4.8cm, height=4cm]{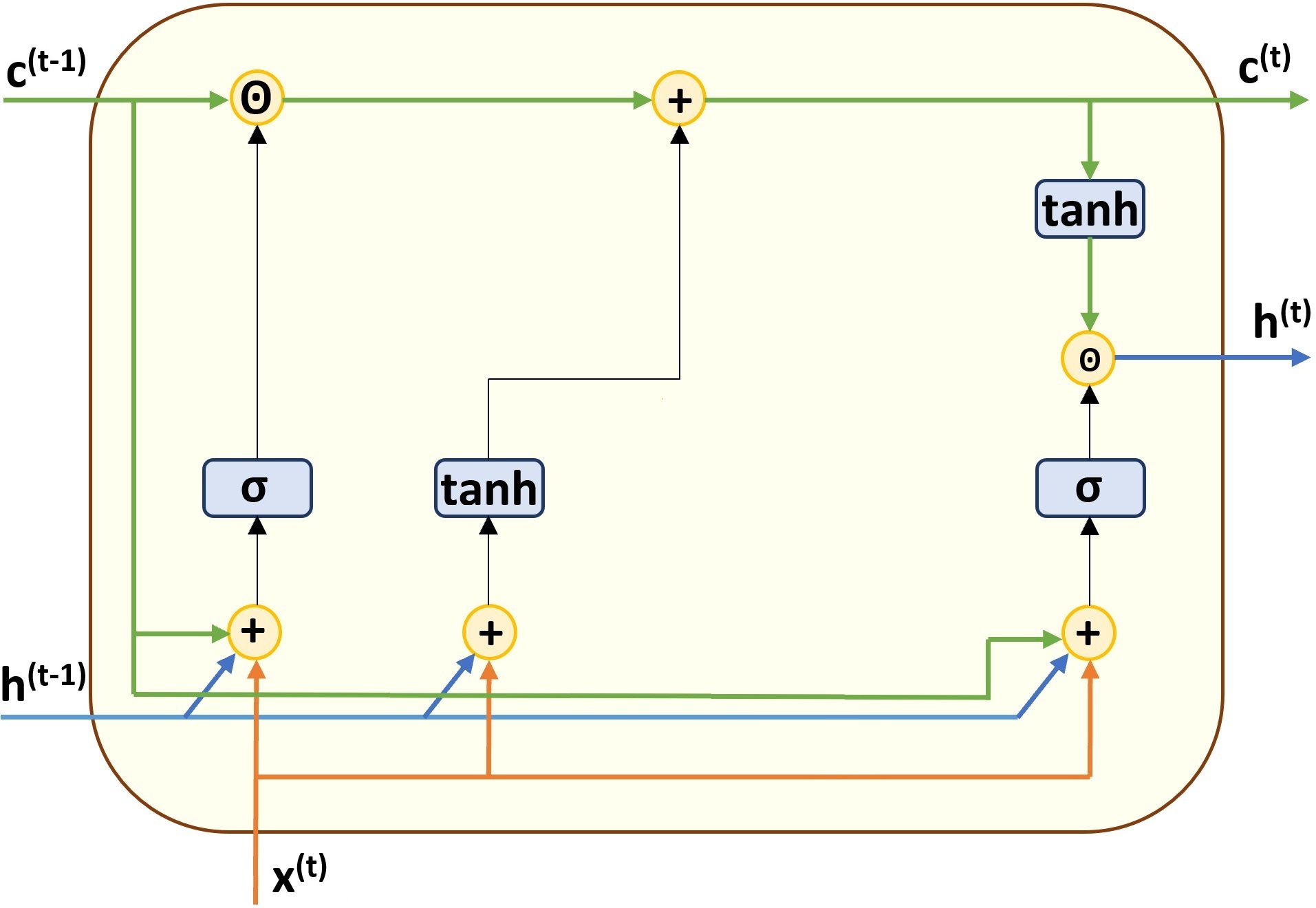}
	\end{minipage} & 
	\parbox{8.5cm}{\begin{align}
		f^{(t)}&=\sigma([W_{fx},U_{fh},W_{fc}] * [x^{(t)}, h^{(t-1)},c^{(t-1)}]  + b_f)\label{eqn49}\\ 
		o^{(t)}&=\sigma([W_{ox},U_{oh},W_{oc}] * [x^{(t)}, h^{(t-1)},c^{(t-1)}]  + b_o)\label{eqn50}\\ 
		g^{(t)}&= \mathrm{tanh}([W_{gx},U_{gh}] *[x^{(t)}, h^{(t-1)}] + b_g)\label{eqn51}\\ 
		c^{(t)}&= f^{(t)}\odot c^{(t-1)} + g^{(t)}\label{eqn52}\\
		h^{(t)}&= \mathrm{tanh}(c^{(t)})\odot o^{(t)}\label{eqn53}
		\end{align}}\\
	\\
	M12&  \begin{minipage}{6cm}
		\includegraphics[width=4.8cm, height=4cm]{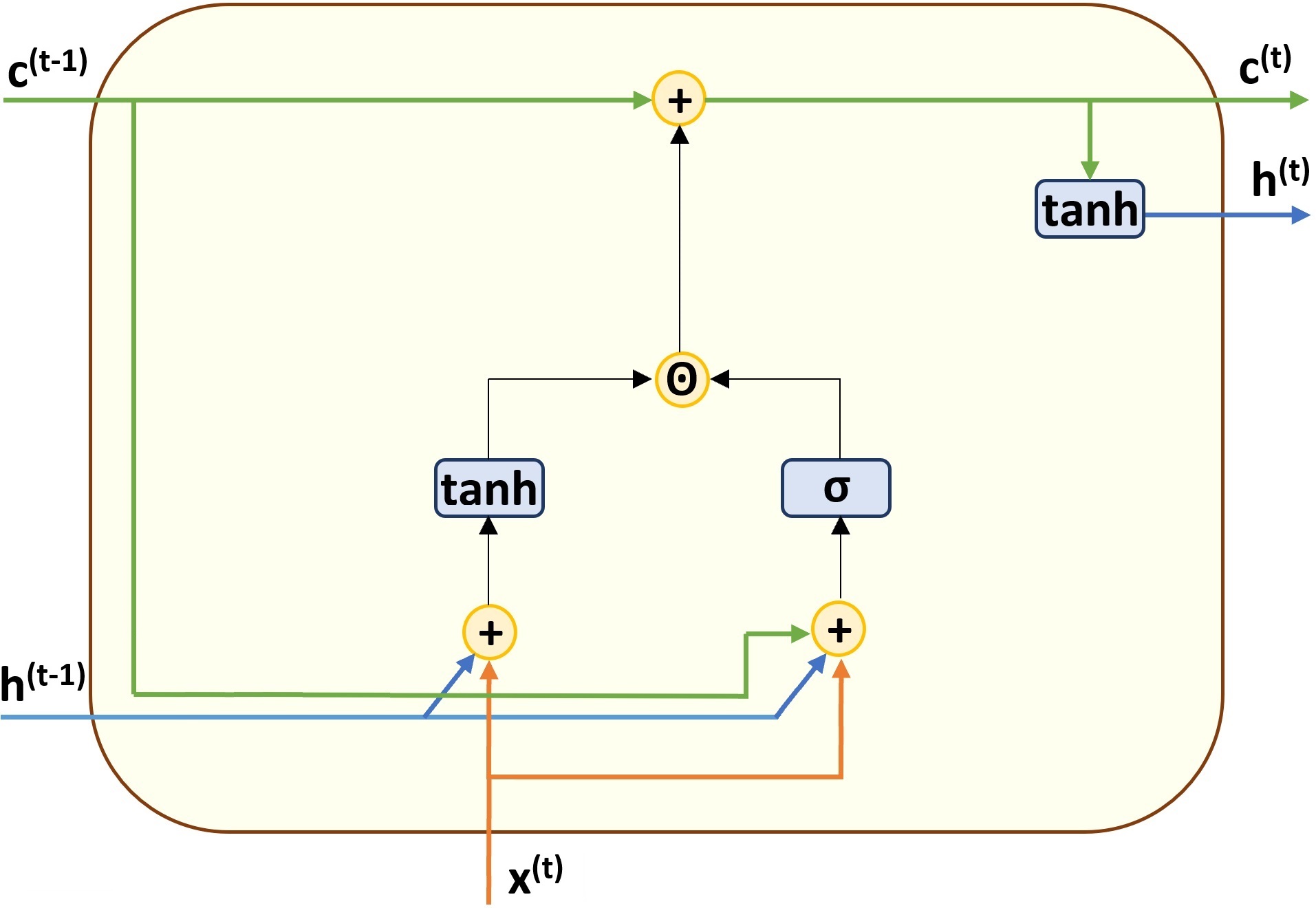}
	\end{minipage} & 
	\parbox{8.5cm}{\begin{align}
		i^{(t)}&= \sigma([W_{ix},U_{ih},W_{ic}]* [x^{(t)}, h^{(t-1)},c^{(t-1)}] + b_i)\label{eqn54}\\ 
		g^{(t)}&=\sigma([W_{gx},U_{gh}] * [x^{(t)}, h^{(t-1)}] + b_g)\label{eqn55}\\ 
		c^{(t)}&= c^{(t-1)} + i^{(t)} \odot g^{(t)}\label{eqn56}\\
		h^{(t)}&= \mathrm{tanh}(c^{(t)})\label{eqn57}
		\end{align}}\\
	\\
	M13&  \begin{minipage}{6cm}
		\includegraphics[width=4.8cm, height=4cm]{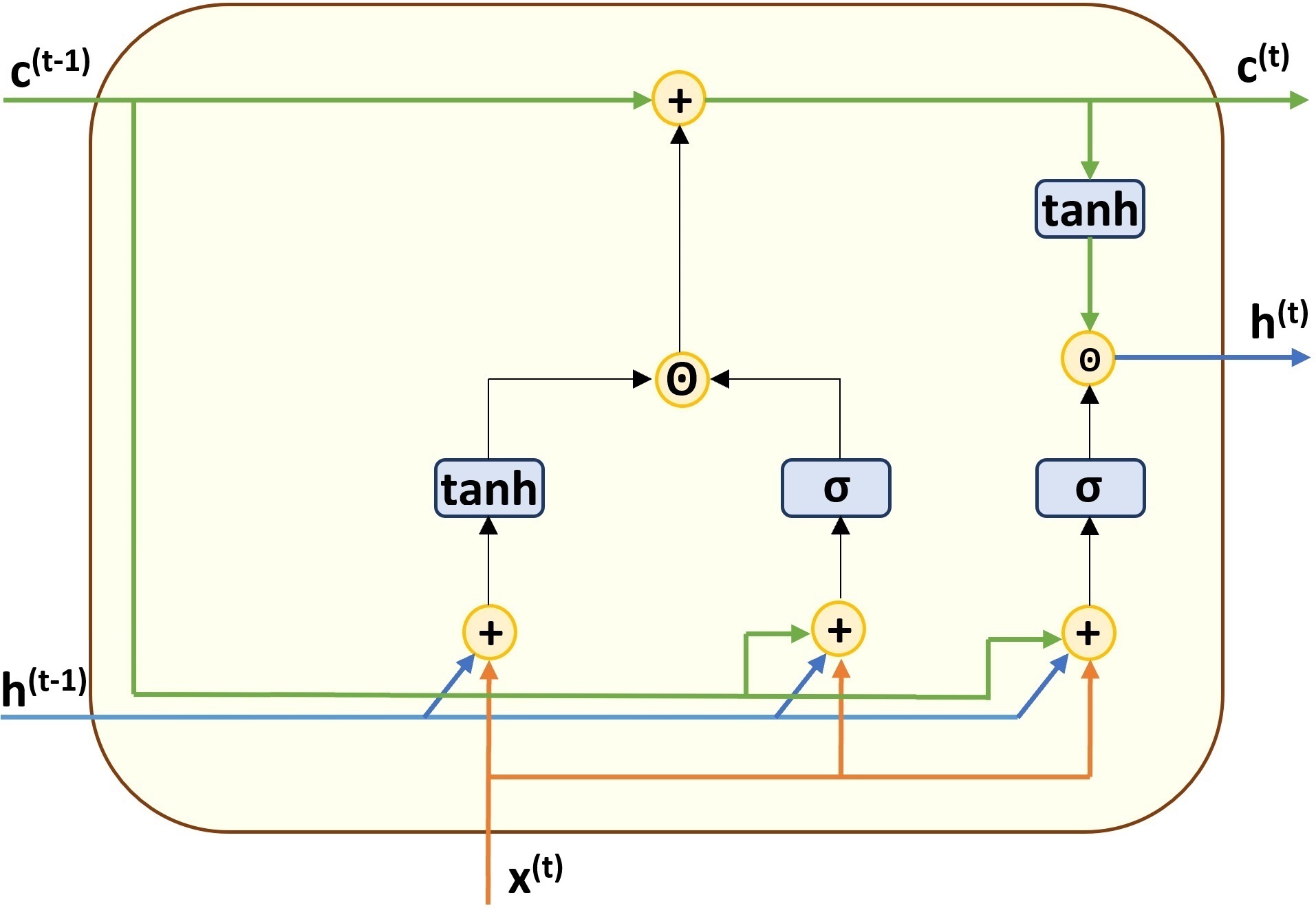}
	\end{minipage} & 
	\parbox{8.5cm}{\begin{align}
		i^{(t)}&= \sigma([W_{ix},U_{ih},W_{ic}]* [x^{(t)}, h^{(t-1)},c^{(t-1)}]  + b_i)\label{eqn58}\\ 
		o^{(t)}&=\sigma([W_{ox},U_{oh},W_{oc}] *[x^{(t)}, h^{(t-1)},c^{(t-1)}]  + b_o)\label{eqn59}\\ 
		g^{(t)}&= \mathrm{tanh}([W_{gx},U_{gh}] *[x^{(t)}, h^{(t-1)}] + b_g)\label{eqn60}\\ 
		c^{(t)}&=c^{(t-1)} + i^{(t)} \odot g^{(t)}\label{eqn61}\\
		h^{(t)}&= \mathrm{tanh}(c^{(t)})\odot o^{(t)}\label{eqn62}
		\end{align}}\\
	\\
	M14&  \begin{minipage}{6cm}
		\includegraphics[width=4.8cm, height=4cm]{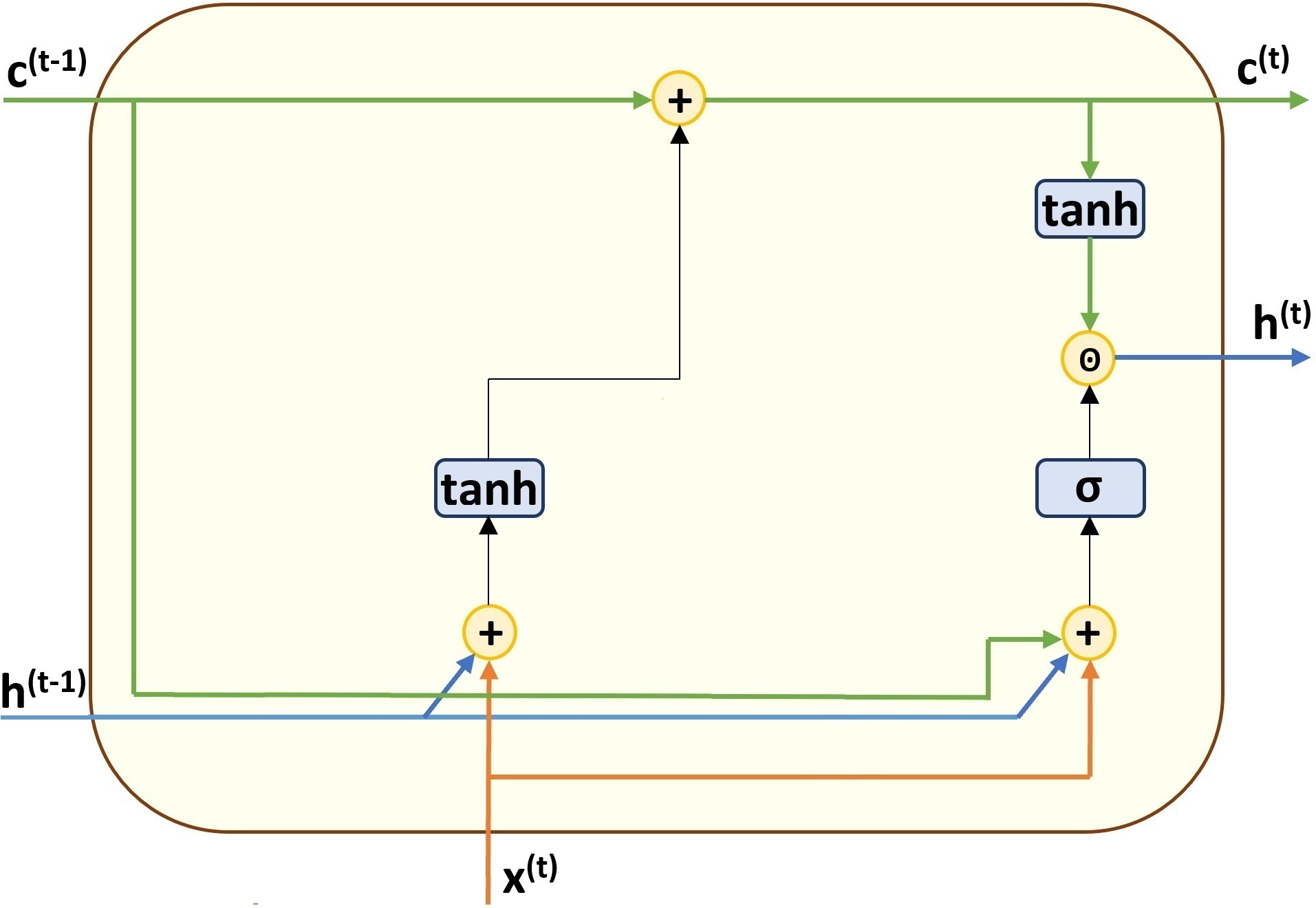}
	\end{minipage} & 
	\parbox{8.5cm}{\begin{align}
		o^{(t)}&=\sigma([W_{ox},U_{oh},W_{oc}] * [x^{(t)}, h^{(t-1)},c^{(t-1)}]  + b_o)\label{eqn63}\\  
		g^{(t)}&= \mathrm{tanh}([W_{gx},U_{gh}] * [x^{(t)}, h^{(t-1)}] + b_g)\label{eqn64}\\ 
		c^{(t)}&= c^{(t-1)} + i^{(t)} \odot g^{(t)}\label{eqn65}\\
		h^{(t)}&= \mathrm{tanh}(c^{(t)})\odot o^{(t)}\label{eqn66}
		\end{align}}\\
	\hline
	\label{non_shared_models}
\end{longtable}

{\scriptsize
	\begin{longtable}{p{1.0cm} p{6.5cm} p{9.6cm}}
		\renewcommand{\arraystretch}{0.7}\\
		\caption{The six multi-function gate variants of convolution LSTM-based models. M15 is an alias for the np-rgcLSTM and M18 is an alias for the rgcLSTM.}\\
		\textbf{Model} &\textbf{\textit{Diagram}}& \textbf{\textit{Formulas}}\\
		\\ \hline
		\\
	M15&  \begin{minipage}{6cm}
		\includegraphics[width=4.8cm, height=4cm]{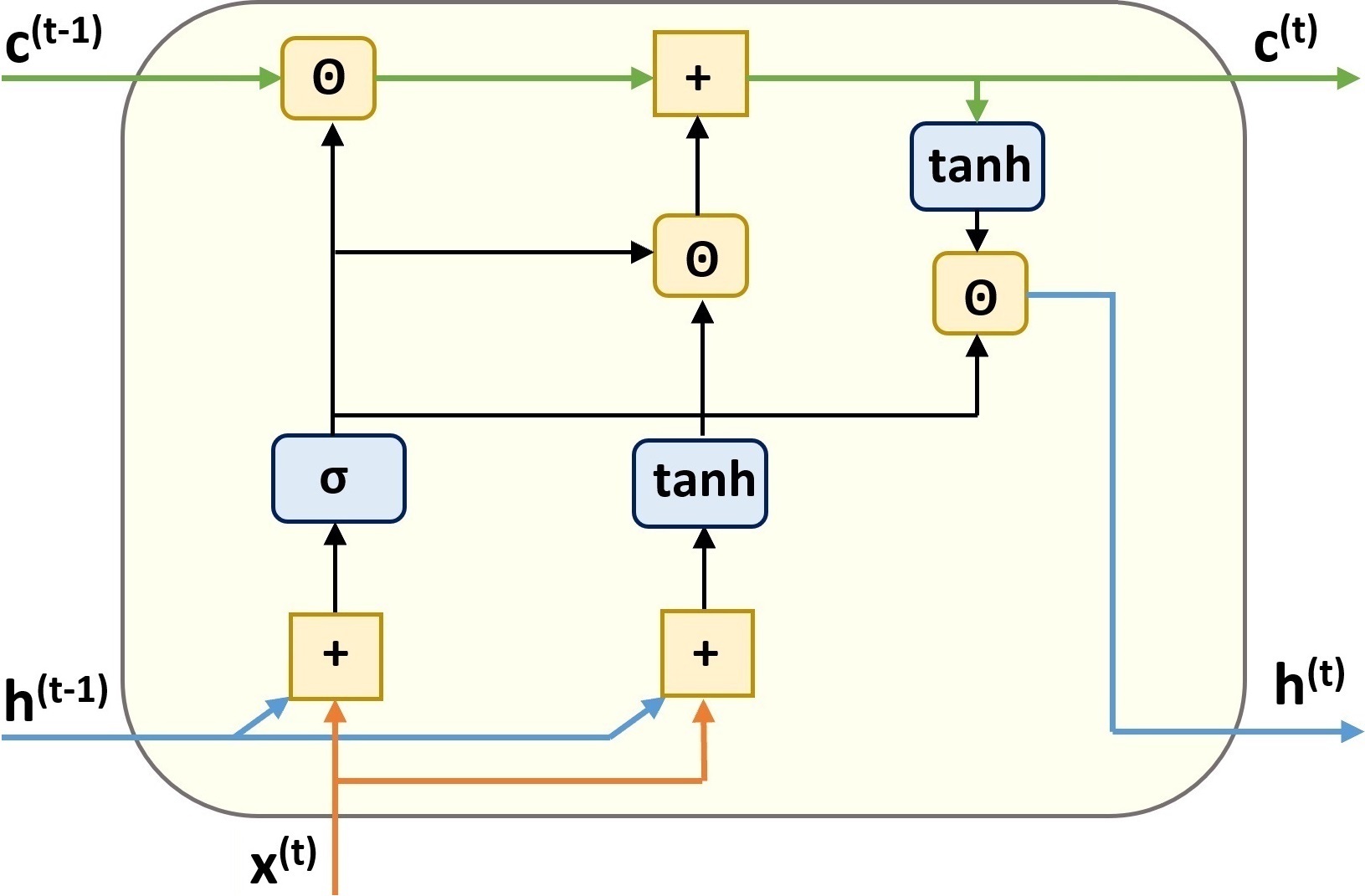}
	\end{minipage} & 
	\parbox{8.5cm}{\begin{align}
		f^{(t)}&=\sigma([W_{fx},U_{fh}] * [ x^{(t)} ,h^{(t-1)}] + b_f)\label{eqn67}\\ 
		g^{(t)}&= tanh([W_{gx},U_{gh}] * [ x^{(t)} ,h^{(t-1)}]+ b_g)\label{eqn68}\\ 
		c^{(t)}&= f^{(t)}\odot c^{(t-1)} + f^{(t)} \odot g^{(t)}\label{eqn69}\\
		h^{(t)}&= tanh(c^{(t)})\odot f^{(t)}\label{eqn70}
		\end{align}}\\
	\\
	M16&  \begin{minipage}{6cm}
		\includegraphics[width=4.8cm, height=4cm]{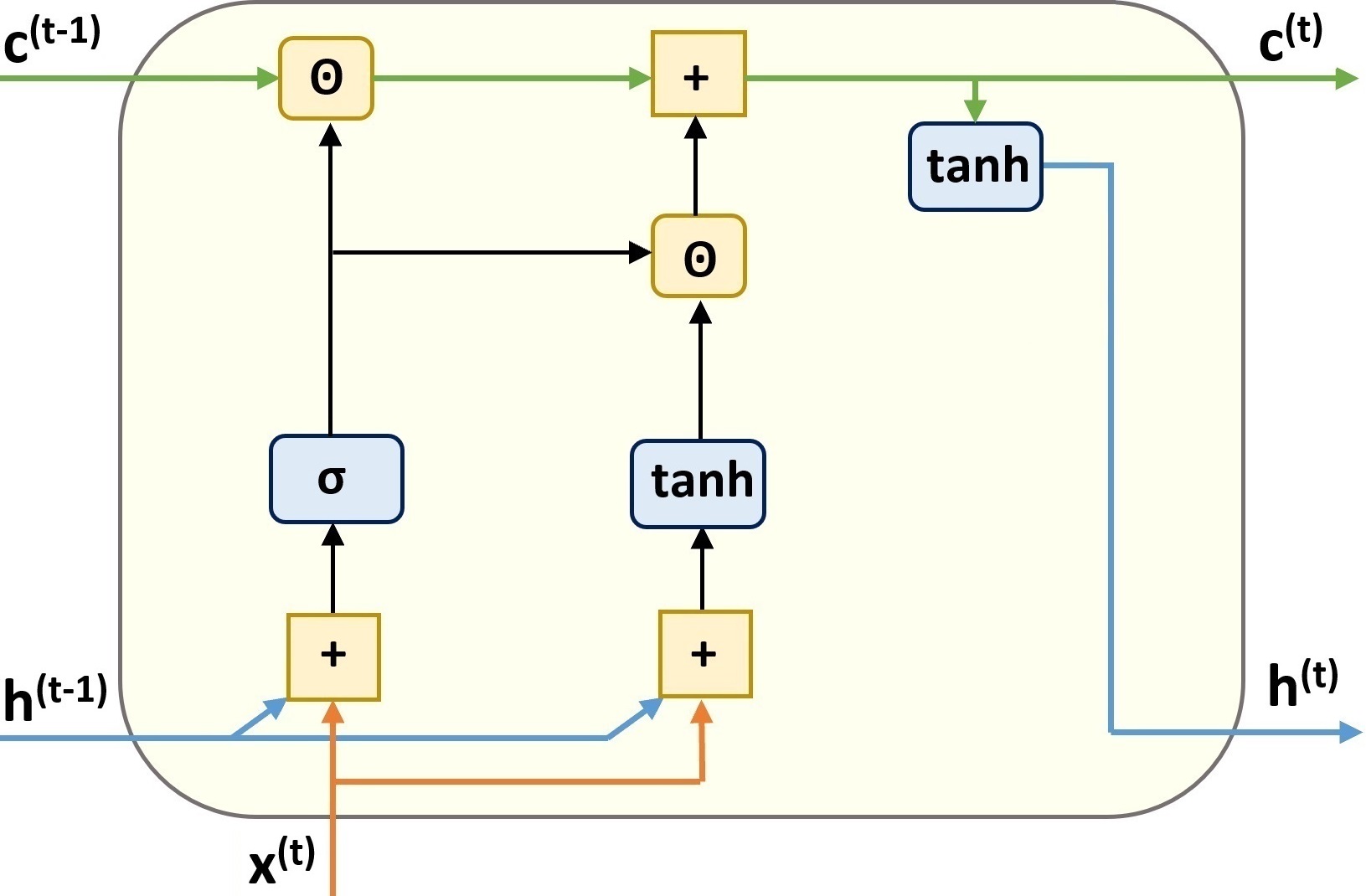}
	\end{minipage} & 
	\parbox{8.5cm}{\begin{align}
		f^{(t)}&=\sigma([W_{fx},U_{fh}] * [ x^{(t)} ,h^{(t-1)}] + b_f)\label{eqn71}\\ 
		g^{(t)}&= tanh([W_{gx},U_{gh}] * [ x^{(t)} ,h^{(t-1)}]+ b_g)\label{eqn72}\\ 
		c^{(t)}&= f^{(t)}\odot c^{(t-1)} + f^{(t)} \odot g^{(t)}\label{eqn73}\\
		h^{(t)}&= tanh(c^{(t)})\label{eqn74}
		\end{align}}\\
	\\
	
	M17&  \begin{minipage}{6cm}
		\includegraphics[width=4.8cm, height=4cm]{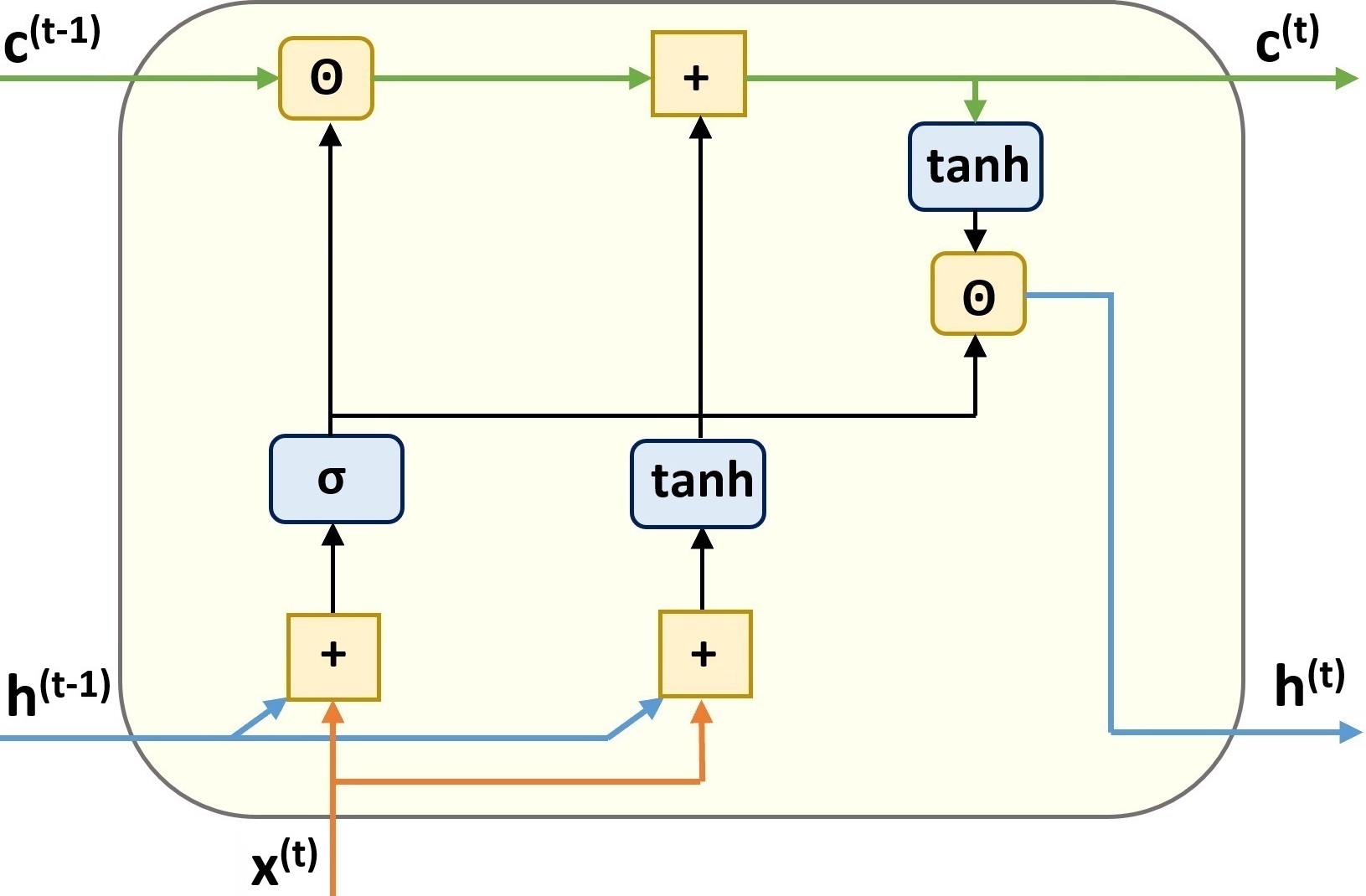}
	\end{minipage} & 
	\parbox{8.5cm}{\begin{align}
		f^{(t)}&=\sigma([W_{fx},U_{fh}] * [ x^{(t)} ,h^{(t-1)}] + b_f)\label{eqn75}\\ 
		g^{(t)}&= tanh([W_{gx},U_{gh}] * [ x^{(t)} ,h^{(t-1)}]+ b_g)\label{eqn76}\\ 
		c^{(t)}&= f^{(t)}\odot c^{(t-1)} + g^{(t)}\label{eqn77}\\
		h^{(t)}&= tanh(c^{(t)})\odot f^{(t)}\label{eqn78}
		\end{align}}\\
	\\
	M18&  \begin{minipage}{6cm}
		\includegraphics[width=4.8cm, height=4cm]{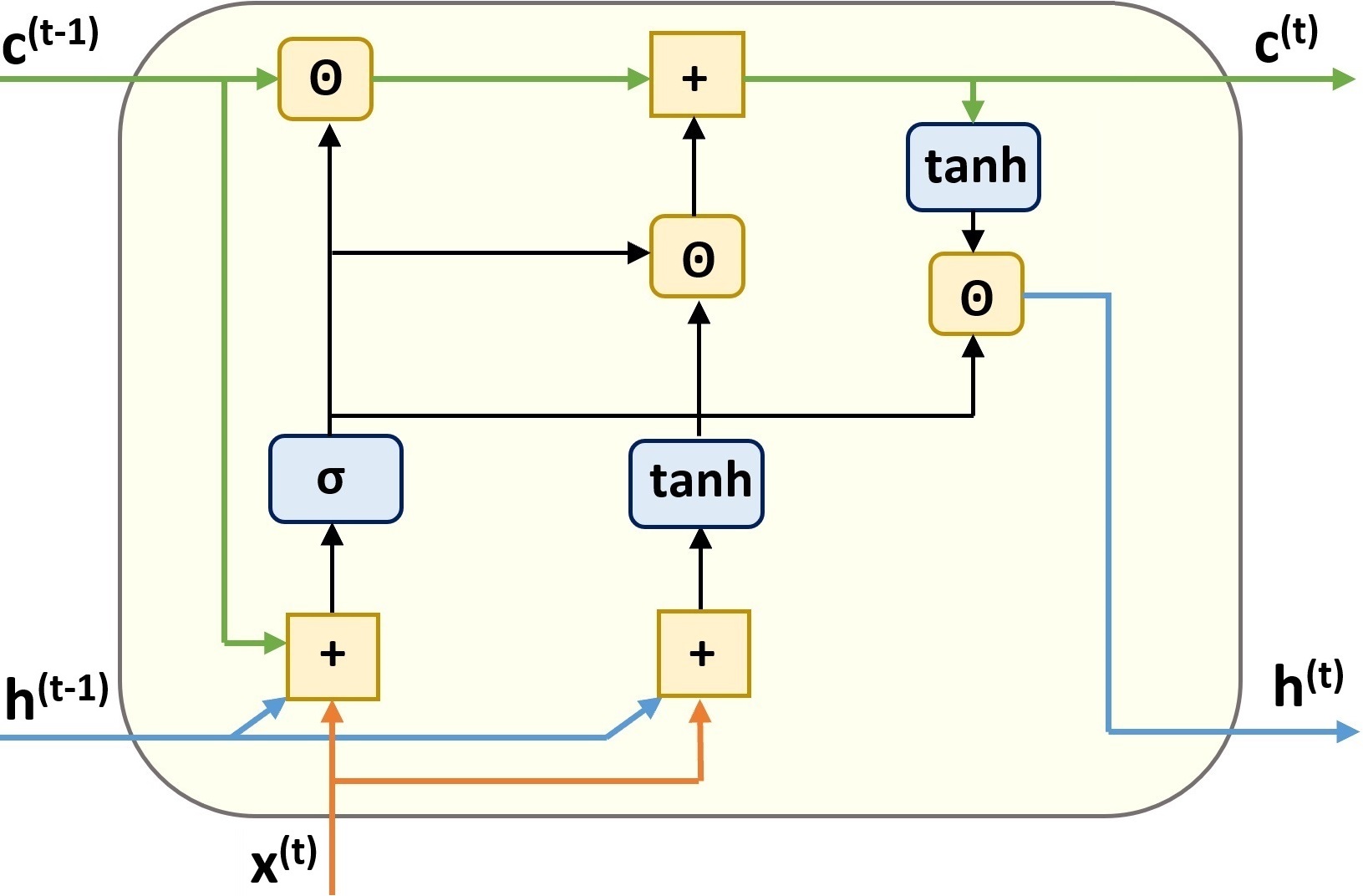}
	\end{minipage} & 
	\parbox{8.5cm}{\begin{align}
		f^{(t)}&=\sigma([W_{fx},U_{fh},W_{fc}] * [ x^{(t)} ,h^{(t-1)},c^{(t-1)}]+ b_f)\label{eqn79}\\ 
		g^{(t)}&= tanh([W_{gx},U_{gh}]*[ x^{(t)} ,h^{(t-1)}]+ b_g)\label{eqn80}\\ 
		c^{(t)}&= f^{(t)}\odot c^{(t-1)} + f^{(t)} \odot g^{(t)}\label{eqn81}\\
		h^{(t)}&= tanh(c^{(t)})\odot f^{(t)}\label{eqn82}
		\end{align}}\\
	\\
	M19&  \begin{minipage}{6cm}
		\includegraphics[width=4.8cm, height=4cm]{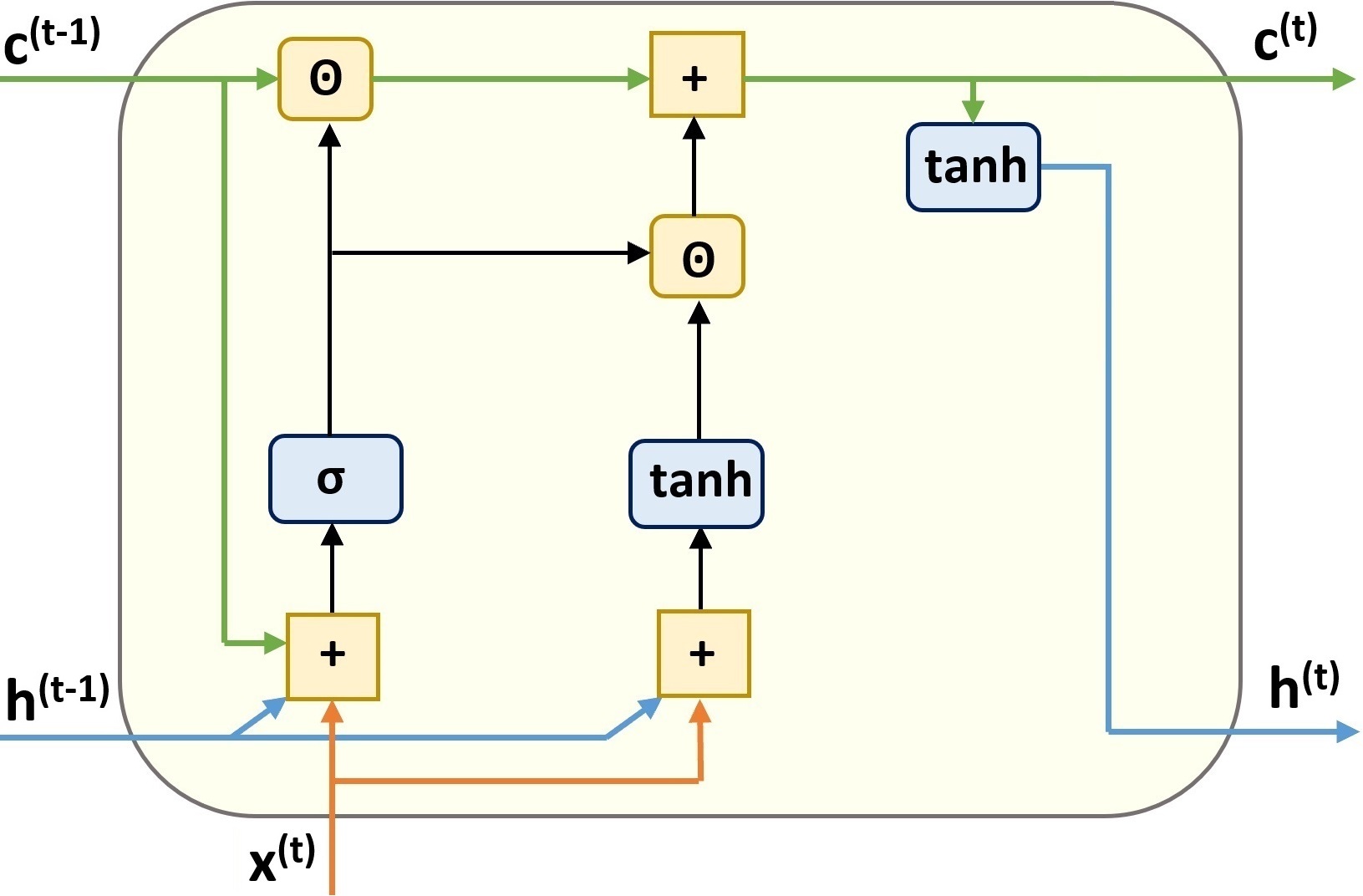}
	\end{minipage} & 
	\parbox{8.5cm}{\begin{align}
		f^{(t)}&=\sigma([W_{fx},U_{fh} , W_{fc}] * [ x^{(t)} ,h^{(t-1)},c^{(t-1)}]+b_f)\label{eqn83}\\ 
		g^{(t)}&= tanh([W_{gx},U_{gh}] * [ x^{(t)} ,h^{(t-1)}]+ b_g)\label{eqn84}\\ 
		c^{(t)}&= f^{(t)}\odot c^{(t-1)} + f^{(t)} \odot g^{(t)}\label{eqn85}\\
		h^{(t)}&= tanh(c^{(t)})\label{eqn86}
		\end{align}}\\
	\\
	
	M20&  \begin{minipage}{6cm}
		\includegraphics[width=4.8cm, height=4cm]{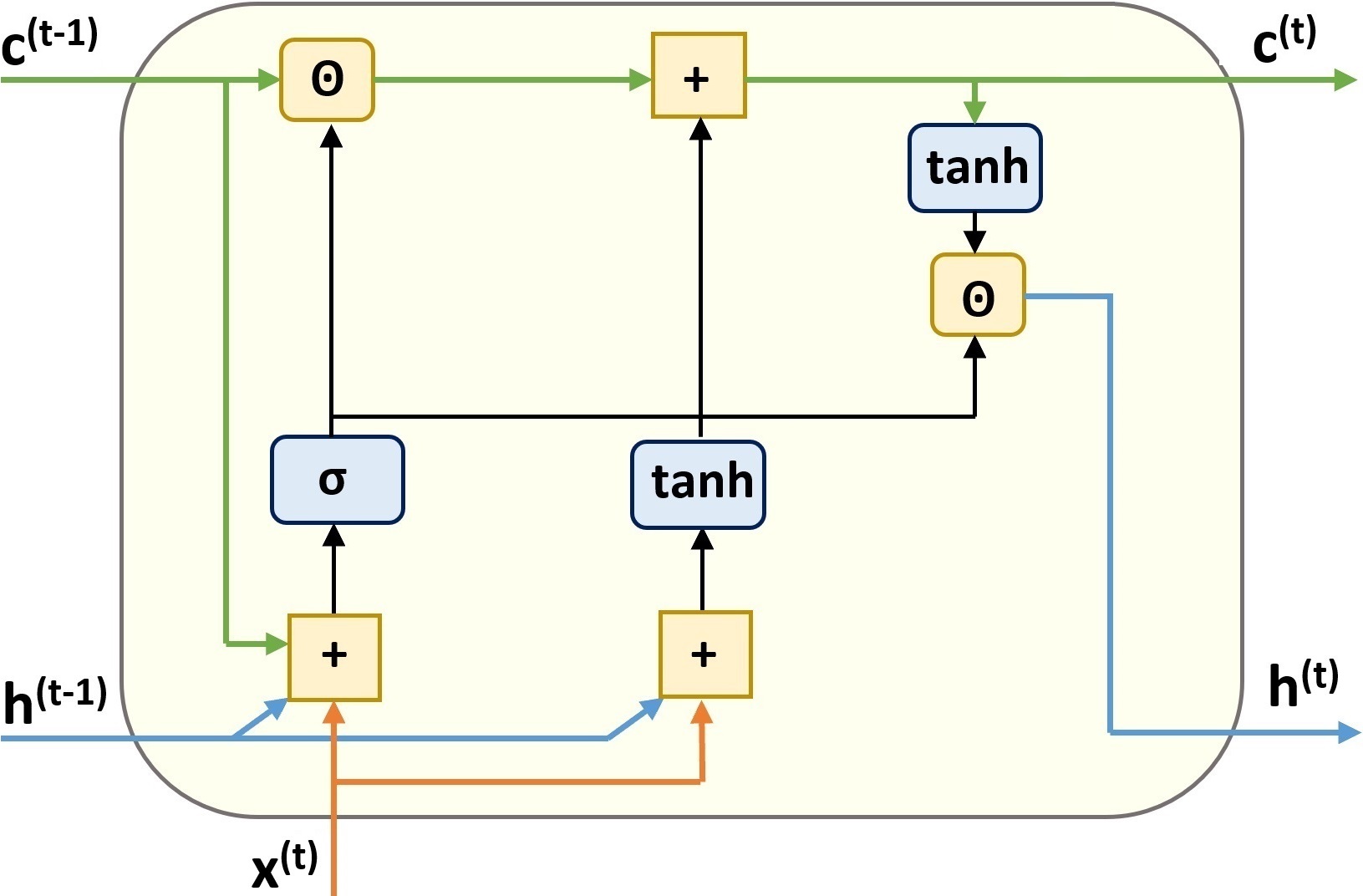}
	\end{minipage} & 
	\parbox{8.5cm}{\begin{align}
		f^{(t)}&=\sigma([W_{fx},U_{fh},W_{fc}] * [ x^{(t)} ,h^{(t-1)},c^{(t-1)}]+ b_f)\label{eqn87}\\ 
		g^{(t)}&= tanh([W_{gx},U_{gh}] * [ x^{(t)} ,h^{(t-1)}]+ b_g)\label{eqn88}\\ 
		c^{(t)}&= f^{(t)}\odot c^{(t-1)} + g^{(t)}\label{eqn89}\\
		h^{(t)}&= tanh(c^{(t)})\odot f^{(t)}\label{eqn90}
		\end{align}}\\
		\\
		\hline

		\label{models_shared}

\end{longtable}}

\bibliography{wileyNJD-AMA}%

\clearpage

\section*{Author Biography}

\begin{biography}{\includegraphics[width=66pt,height=86pt,]{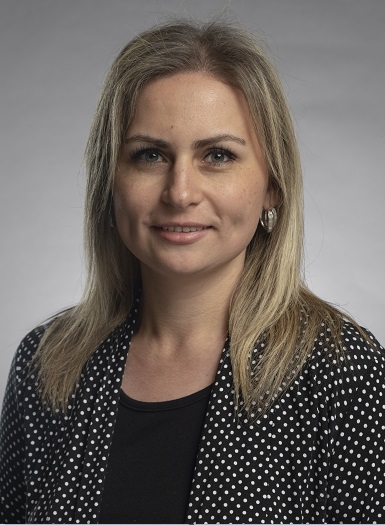}}{\textbf{Nelly ELsayed} is an assistant professor at the School of Information Technology, University of Cincinnati. She received her Bachelor degree in Computer Science from Alexandria University in 2010 and she was ranked the First on Computer Science Class. She received two Master degrees, the first in Computer Science in 2014 at Alexandria University and the second in Computer Engineering in 2017  from the University of Louisiana at Lafayette. She received her Ph.D. in Computer Engineering from the University of Louisiana at Lafayette. She is an honor member in the national society of leadership and the honor society Phi Kappa Phi. Her major research interests are in machine learning, artificial intelligence, convolutional recurrent neural networks, bio-inspired computations, information technology, data science, and data analysis.}
\end{biography}
\begin{biography}{\includegraphics[width=66pt,height=86pt,]{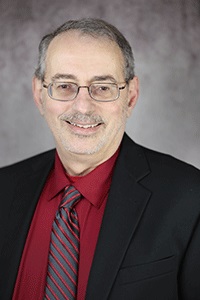}}{\textbf{Anthony S. Maida} is associate professor,
Alfred and Helen M.\ Lamson Endowed Professor in computer science,  
and graduate coordinator for computer science and computer engineering at School of Computing and Informatics, University of Louisiana at Lafayette. He received this BA in Mathematics in 1973, Ph.D. in Psychology in 1980, and Masters Degree in Computer Science in 1981, all from the University of Buffalo. He has done two Post Doctoral degrees at Brown University and the University of California, Berkeley. He was a member of the computer science faculty at the Penn State University from 1984 through 1991. He has been a member of a Center for Advanced Computer Studies and School of Computing and informatics at the University of Louisiana at Lafayette from 1991 to the present. His research interests are: intelligent systems, neural networks, recurrent neural networks and brain simulation.}
\end{biography}

\begin{biography}{\includegraphics[width=66pt,height=86pt,]{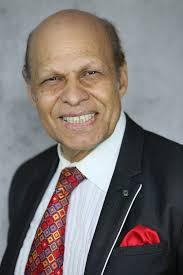}}{\textbf{Magdy Bayoumi} is department head and professor at the Electrical Engineering Department, University of Louisiana at Lafayette. He was the Director of CACS, 1997 to 2013 and Department Head of the Computer Science Department, 2000-2011. Dr. Bayoumi has been a faculty member in CACS since 1985. He received B.Sc. and M.Sc. degrees in Electrical Engineering from Cairo University, Egypt; M.Sc. degree in Computer Engineering from Washington University, St. Louis; and Ph.D. degree in Electrical Engineering from the University of Windsor, Canada. He is on the IEEE Fellow Committee and he was on the IEEE CS Fellow Committee. His research interests are: technology, data processing, management, internet of things, energy and security.}
\end{biography}

\end{document}